\documentclass[preprint,12pt]{elsarticle}

\usepackage{amssymb}
\usepackage{graphicx} 
\usepackage{amsmath, amssymb} 
\usepackage{float} 
\usepackage{hyperref}
\usepackage{setspace}
\usepackage{caption}
\usepackage{xcolor}
\usepackage{hypcap}
\usepackage{graphicx}
\usepackage{subcaption}

\usepackage{siunitx}
\usepackage{array}
\usepackage{booktabs}
\usepackage{tabularx}
\definecolor{revblue}{RGB}{0,90,200}
\newcommand{\new}[1]{\textcolor{black}{#1}}
\newenvironment{newpar}{\color{black}}{}

\journal{Reliability Engineering \& System Safety}

\begin{document}

\begin{frontmatter}

\title{WaveGraphNet: Physics-Consistent Guided-Wave Damage Localization through Coupled Inverse--Forward Graph Learning}

\author[inst1]{Vinay Sharma\fnref{equal}}
\author[inst1]{Aditya Bharade\fnref{equal}}
\author[inst1]{Olga Fink\corref{cor1}}

\affiliation[inst1]{organization={EPFL, Intelligent Maintenance and Operations Systems},
            city={Lausanne},
            country={Switzerland}}

\fntext[equal]{These authors contributed equally to this work.}
\cortext[cor1]{Corresponding author}

\begin{abstract}
Guided-wave structural health monitoring enables damage localization in composite plates using sparse networks of bonded piezoelectric transducers. However, supervised localization remains weakly constrained when measurements are available from only a limited set of damage locations. Because exhaustive spatial coverage is impractical, models trained at observed locations may generalize poorly to unseen regions.

We propose \textit{WaveGraphNet}, an inverse--forward graph-learning framework for guided-wave damage localization in carbon-fiber-reinforced polymer (CFRP) plates. The sensing network is represented as a graph, with transducers as nodes and measured pitch--catch paths as edges. The inverse branch uses order-invariant message passing and a geometry-constrained decoder to map path-wise energy deviations to a damage coordinate. An independently trained forward branch predicts the energy-deviation pattern associated with a candidate coordinate. At test time, gradients propagated through the forward branch update only the inverse-predicted coordinate to reduce the discrepancy between the measured and predicted response patterns.

The framework is evaluated on the OGW-1 benchmark using three spatial hold-out splits in which complete damage regions are excluded from training. Comparisons include non-graph and graph-learning baselines together with the training-free Reconstruction Algorithm for Probabilistic Inspection of Damage (RAPID). Under the validation-selected settings, refined \textit{WaveGraphNet} achieves the lowest reported test mean absolute error on all three splits. Refinement reduces the standalone localization error by \(23.6\%\)--\(67.9\%\), with an average relative reduction of \(50.4\%\). These results demonstrate that learned forward-response consistency can improve guided-wave localization under limited spatial training coverage.
\end{abstract}

\begin{highlights}
\item Sparse guided-wave damage localization is formulated as a graph-based regression problem.
\item Coupled inverse-forward architecture localizes damages consistent with energy deviation.
\item An independently trained forward model refines coordinates at test time.
\item Refined model outperforms all evaluated baselines, including RAPID.
\end{highlights}


\end{frontmatter}



\section{Introduction}
\label{sec:introduction}

Engineering structures require monitoring strategies that can detect and localize barely visible or internal damage before it propagates to a critical state \cite{guemes2020composite_shm_review}. Composite materials, and in particular Carbon Fiber Reinforced Polymers (CFRP), present unique challenges in this context due to their anisotropy and complex damage mechanisms. For structural health monitoring (SHM) to be practically useful, however, methods must bridge the persistent gap between controlled laboratory demonstrations and reliable field deployment \cite{cawley2018closing_gap}. Guided ultrasonic waves are attractive in this context because they can interrogate relatively large plate-like regions using sparse networks of surface-bonded piezoelectric transducers \cite{ricci2022guided_waves_composites_review,capineri2021guided_wave_sensors_review,philibert2022lamb_waves_aeronautics}. Such sensing layouts are especially relevant for composite structures, including CFRP components, and recent reviews emphasize the broader importance of guided waves for SHM and long-range nondestructive testing \cite{cawley2024guided_waves_ndt_shm,tanveer2024guided_waves_laminated_composites}. The central challenge, however, is that damage location is never observed directly; it must be inferred from how a defect perturbs multiple actuator-receiver paths across a sparse sensing network.

This inverse mapping is difficult even under controlled experimental conditions. Composite anisotropy, multimodal propagation, dispersion, boundary reflections, and limited path coverage all reduce identifiability, especially when different damage locations produce similar scattered responses \cite{ricci2022guided_waves_composites_review,philibert2022lamb_waves_aeronautics}. The problem becomes more severe when only a small number of labeled damage positions is available for model development, which is the realistic regime for many SHM applications. A localization model may therefore appear accurate on the sampled damage grid while remaining unreliable in unsampled regions of the structure, where it must effectively extrapolate beyond the locations seen during training. In SHM, this is not merely an evaluation detail but a deployment issue: practical systems must remain reliable beyond the specific defect locations observed during data collection \cite{cawley2018closing_gap}.

\begin{newpar}
Addressing this difficult inverse mapping has motivated learning-based guided-wave localization methods that extract damage-sensitive information from measured signals \cite{sattarifar2022ml_guided_wave_shm_review}. Existing approaches include image-based signal representations \cite{liao2023gramian_guided_wave_localization} and artificial neural networks for damage localization and severity estimation \cite{gao2024lamb_wave_modular_ann,azad2024lamb_wave_localization_severity}. 
These approaches process the measurements as arrays, sequences or images. However, a sparse pitch--catch network is instead inherently relational: each measured quantity belongs to a specific transducer pair, and localization requires the joint interpretation of evidence distributed across multiple propagation paths. Graph models provide a natural representation of this topology and have already been investigated for guided-wave damage assessment \cite{Wang2022GiG,sun2024physics_augmented_stgcn_guided_waves}.

Even with a graph representation, coordinate regression remains weakly constrained when measurements from only a limited set of damage locations are available for training. 
Distinct locations may produce similar path-wise perturbation patterns, and a prediction can be numerically plausible without being consistent with the response measured across the full sensing network. Moreover, representing one physical pitch--catch measurement as two independently parameterized directed edges can introduce an artificial dependence on the arbitrary ordering of the two transducers in the graph.
These challenges expose two requirements for robust localization from sparse training data: the representation of each sensing path should be invariant to transducer ordering, and the inferred damage location should not only be close to labeled coordinates but also explain the response observed across the sensing network. \new{The latter requirement is a form of observational bias, which has been shown to be an effective way of constraining learning when training data are scarce~\cite{fink2026physics2}.}

\textit{WaveGraphNet} is designed around these two requirements. We propose an inverse--forward graph framework that combines direct coordinate inference with measurement-based response consistency. The inverse branch uses order-invariant message passing and a geometry constrained decoder to map graph-structured measurements of path-wise energy deviation to a defect coordinate. The complementary forward branch learns the reverse mapping: from a candidate coordinate and the known sensing geometry to the corresponding path-wise energy-deviation pattern. At test time, the inverse prediction initializes the candidate coordinate supplied to the forward branch. The gradients propagated through the forward branch are used to refine the predicted coordinate by minimizing the mismatch between this reconstructed energy pattern and the measured one. In this way, localization is guided not only by the coordinate regressor but also by whether the inferred location can reproduce the spatial redistribution of wave-response energy observed across the sensing network. The forward branch is a learned response-consistency surrogate, not a first-principles wave-propagation solver; its role is to provide a differentiable, measurement-based consistency signal rather than to enforce dispersion, anisotropy, or time-of-flight equations.
\end{newpar}

The main contributions of this work are as follows:
\begin{itemize}
    \item \new{We formulate sparse guided-wave localization as graph regression, with the 12 transducers represented as nodes and the 66 measured transducer pairs represented as bi-directional edges. The formulation treats the two transducers of each pair symmetrically, so exchanging their order does not change the graph representation or localization prediction.}
    \item \new{We introduce a geometry-constrained inverse graph model that aggregates path-level localization evidence through a voting mechanism to infer a defect coordinate for damaged samples. For undamaged samples, the regression target is defined as an explicit coordinate outside the sensed-domain.}
    \item \new{We introduce an independently trained forward response surrogate that provides a measurement-based consistency constraint for a labeled damage coordinate. At test time, the inverse-predicted damage coordinate is refined by minimizing the discrepancy between the measured and reconstructed path-wise response patterns, without updating any network weights. This refinement reduces the mean localization error by an average of 50\% across the three evaluated spatial splits.}
    \item \new{We systematically evaluate generalization to unseen damage locations using three held-out-region splits of the Open Guided Waves OGW-1 benchmark, including a challenging cross-region setting, and benchmark the proposed approach against both learned localization models and the training-free RAPID elliptical-imaging method.}
\end{itemize}

\new{The remainder of the paper is organized as follows. Section~\ref{sec:rel_work} reviews learning-based localization, graph-based SHM, learned consistency mechanisms, and classical imaging. Section~\ref{sec:problem_formulation} defines the localization problem and the path-wise consistency signal. Section~\ref{sec:method} presents the inverse and forward modules, their independent training, and test-time refinement. Section~\ref{sec:case_study} describes the OGW-1 case study and spatial evaluation protocol. Sections~\ref{sec:implementation_details} and \ref{sec:results} present the comparative setup and results, and Section~\ref{sec:discussion} concludes the paper.}

\section{Related Work}
\label{sec:rel_work}

\begin{newpar}

Existing approaches to guided-wave damage localization range from data-driven mappings of measured responses to methods that incorporate sensing geometry or physical consistency. We review four areas most relevant to this work: learning-based guided-wave localization, graph-based modeling for SHM, forward consistency in inverse problems, and classical training-free imaging.

\textbf{Learning-based guided-wave localization.}
A growing body of work has investigated how damage location can be inferred directly from measured guided-wave responses. These responses have been represented as time series, engineered feature vectors, and image-like transformations. Gramian-angular-field representations combined with convolutional networks have been used for localization~\cite{liao2023gramian_guided_wave_localization}; engineered envelope features have been paired with neural models for composite structures~\cite{gao2024lamb_wave_modular_ann}; and deep networks have addressed joint localization and severity estimation from Lamb-wave signals~\cite{azad2024lamb_wave_localization_severity}. Unified pipelines have also combined detection, localization, and quantification ~\cite{lomazzi2023unified_damage_detection_localization_quantification}. Unsupervised formulations have likewise been used to detect and localize structural damage without labelled damage states~\cite{liu2024structural}. These studies demonstrate that guided-wave measurements contain rich information about damage location. However, conventional array- and sequence-based representations are not naturally structured around the pairwise sensing geometry of pitch–catch measurements, motivating representations that associate each measured response directly with its corresponding propagation path.

\textbf{Graph-based modeling for SHM.}
Graphs provide a natural representation of sensor layout, structural connectivity, and spatial dependence directly, and have consequently been adopted in SHM and graph-signal-processing frameworks \cite{bloemheuvel2021graph_shm_framework,cheema2024graph_signal_processing_shm, niresi2026timevertex}. Graph models have also been studied for aerospace damage localization \cite{delpriore2025gnn_aerospace_damage_localization} and for guided-wave sensing through graph-in-graph and physics-augmented spatio-temporal architectures \cite{Wang2022GiG,sun2024physics_augmented_stgcn_guided_waves}. These studies demonstrate the value of graph representations for incorporating sensing geometry, but they do not specifically address the ambiguity introduced by representing a single physical pitch–catch path through directed graph relations, nor do they use the inferred coordinate to reconstruct the measured network-wide response as a consistency check. Our work builds on graph-based representations while targeting these two aspects: each measured path is represented by a pair of directed edges tied to a shared, order-invariant embedding that contributes a geometry-constrained coordinate vote, rather than by two independently learnable directed relations. Its methodological distinction is the use of a separately trained forward surrogate to refine coordinates under spatial hold-out conditions.

\textbf{Forward consistency in inverse problems.}
Physics-consistent guided-wave methods have incorporated simulated responses from reduced-order spectral finite-element models \cite{rautela2021model_assisted_guided_wave}, combined models of differing fidelity for probabilistic damage diagnosis~\cite{miele2023multifidelity} and enforced time-of-flight consistency to suppress boundary reflections before inference \cite{song2024physics_guided_lamb_wave}. More broadly, forward models and consistency objectives are used to reduce ambiguity in inverse problems \cite{ongie2020inverse_imaging_review,raissi2019physics,niresi2025rins,fink2026physics}. \new{Such measurement-driven constraints correspond to an observational bias in physics-informed learning~\cite{fink2026physics2}.}
\textit{WaveGraphNet} adopts this general principle without relying on a first-principles wave-propagation model. Instead, its forward branch learns the mapping from candidate damage coordinates and sensing geometry to path-wise response deviations directly from data. It therefore provides a learned response-consistency constraint: a candidate location is supported when it can reproduce the redistribution pattern observed across the sensing network.

\textbf{Classical training-free imaging.}
Probabilistic imaging methods such as the Reconstruction Algorithm for Probabilistic Inspection of Damage (RAPID) combine path-wise damage indices using elliptical geometric kernels. They require no localization training data and therefore provide an important reference when learned models are evaluated outside the spatial support of their training locations. RAPID was first proposed by Gao et al.~\cite{gao2005rapid} and subsequently demonstrated for guided-wave inspection of aircraft structures and pipes \cite{zhao2007rapid,vanvelsor2007rapid}. It has also been applied to impact-damage localization in CFRP plates \cite{hettler2016rapid}. Recent work has applied this class of methods to open guided-wave benchmarks \cite{song2025unsupervised}. Unlike learned localization approaches, these methods rely directly on sensing geometry and measured path-wise changes, avoiding dependence on the spatial distribution of labeled damage locations. This makes them a particularly relevant point of comparison for assessing whether learned models can generalize beyond the damage locations represented during training.

\end{newpar}

\section{Problem Formulation}
\label{sec:problem_formulation}

We consider a guided-wave SHM system for damage localization in a composite plate, comprising \(N\) bonded piezoelectric transducers
 operated in pitch--catch mode. In each measurement, one transducer excites a guided wave and another records the resulting response. The
transducers can serve as both actuators and receivers. Let
\(V=\{1,\ldots,N\}\) denote the transducer set, and let
\(\mathbf{r}_i\in\mathbb{R}^2\) be the in-plane coordinate of transducer
\(i\).

The acquisition provides a set of measured pitch--catch channels, denoted by
\(\mathcal{P}\). Each element of \(\mathcal{P}\) corresponds to one measured
actuator--receiver transmission between two distinct transducers. We refer
to each such measured channel as a \emph{propagation path}. Along this
channel, the emitted guided wave travels through the plate, interacts with
the structure, and may be perturbed by damage before being recorded at the
receiver.

For the OGW-1 dataset considered in this study, a single pitch–catch measurement is available for each pair of distinct transducers; reversing the ordering of the transducer pair does not define an additional measurement. We therefore define \(\mathcal{P}\) as the set of measured propagation paths, with each physical transducer pair represented only once.
\new{For \(N=12\), this gives
\(|\mathcal{P}|=\binom{12}{2}=66\) propagation paths per sample.}
For notational convenience, each path is denoted by
\((i,j)\in\mathcal{P}\), where \(i\) and \(j\) identify the two transducers associated with the measurement; the ordering is purely notational and does not define a propagation direction.

For each measured propagation path \((i,j)\in\mathcal{P}\), a discrete
response
\begin{equation}
s_{ij}(t), \qquad t=1,\ldots,T,
\end{equation}
is recorded over \(T\) time samples.

\begin{newpar}
The localization target is defined in normalized plate coordinates. For a
damaged sample, the target is the defect coordinate
\(\mathbf{p}=[x_d,y_d]^\top\) in the admissible plate domain
\(\Omega=[0,1]^2\). Undamaged samples are assigned a fixed reference
coordinate outside this domain. The unified target is therefore
\begin{equation}
\mathbf{p}_{\mathrm{tar}}
=
\begin{cases}
[x_d,y_d]^\top \in \Omega, & \text{damaged sample},\\[2mm]
\mathbf{p}_{\mathrm{ud}}=[-0.5,-0.5]^\top,
& \text{undamaged sample}.
\end{cases}
\label{eq:unified_target}
\end{equation}
The objective here is to infer \(\mathbf{p}_{\mathrm{tar}}\) from the collection
of measured path-wise responses.  We focus on generalization to unseen parts of the structure of the structure, which we refer to as spatial extrapolation: damage must be localized at locations where no damaged training samples were collected.
\end{newpar}

\subsection{Damage-sensitive signal representation}
\label{sec:signal_representation}

\begin{newpar}
The raw signals contain strong direct arrivals, boundary reflections, and measurement noise that can obscure defect-induced effects. To isolate damage-sensitive perturbations, a pristine reference is estimated for each propagation path and subtracted from the measured response. Let \(\bar{s}_{ij}(t)\) denote the mean pristine signal for path \((i,j)\). The differential signal is given by
\begin{equation}
\delta s_{ij}(t)=s_{ij}(t)-\bar{s}_{ij}(t).
\label{eq:differential_signal}
\end{equation}
This pristine-reference subtraction suppresses the dominant response of the undamaged structure while retaining damage-induced changes in the measured signal.

Each differential signal is transformed into the frequency domain over a
selected band \(\mathcal{F}=\{f_1,\ldots,f_K\}\). Its absolute amplitude is
extracted at each frequency bin and normalized using statistics computed
exclusively from the training partition, giving
\(\tilde{A}_{ij}(f_k)\). Let \(\bar{A}_{ij}(f_k)\) denote the corresponding
mean absolute normalized amplitude over pristine training samples. We define
the path-wise response-deviation index
\begin{equation}
\Delta E_{ij}
=
\frac{1}{E_{\max}}
\max\!\left(
\frac{1}{K}
\sum_{k=1}^{K}
\left(
\tilde{A}_{ij}(f_k)-\bar{A}_{ij}(f_k)
\right),
0
\right)
\label{eq:deltaE}
\end{equation}
where \(E_{\max}\) is the maximum deviation observed over all training
samples and measured paths. The non-negativity clamp retains only positive
mean amplitude deviations, and normalization maps the
quantity to \([0,1]\). Although denoted by \(\Delta E_{ij}\), it is a
normalized spectral-amplitude-deviation index rather than a direct
measurement of mechanical wave energy.

Concatenating the values over all measured paths gives
\begin{equation}
\Delta\mathbf{E}
=
\left[
\Delta E_{ij}
\right]_{(i,j)\in\mathcal{P}}
\in [0,1]^{|\mathcal{P}|}.
\label{eq:deltaE_vector}
\end{equation}
Where, for the OGW-1 configuration, \(\Delta\mathbf{E}\in\mathbb{R}^{66}\) corresponds to one scalar value for 66 measured propagation paths.
\end{newpar}

\subsection{Graph representation of one measurement sample}
\label{sec:graph_representation_one_meas}

\begin{newpar}
Damage localization is reflected not by the response of a single propagation path, but by the joint pattern of perturbations across the sensing network. We therefore represent each measurement sample as an observation graph
\begin{equation}
\mathcal{G}_{\mathrm{obs}}
=
\left(
V,\,
\mathcal{P},\,
\{\mathbf{r}_i\}_{i\in V},\,
\{\Delta E_{ij}\}_{(i,j)\in\mathcal{P}}
\right).
\label{eq:observation_graph}
\end{equation}
The nodes correspond to transducers with known coordinates, while each measured
unordered transducer pair defines a graph edge carrying its
measured energy-deviation value. The geometry of each propagation path, including its length
\(\|\mathbf{r}_j-\mathbf{r}_i\|_2\), is determined directly from the known
transducer coordinates. 
\end{newpar}

\subsection{\new{Weakly constrained localization and response consistency}}
\label{sec:localization_consistency}

\begin{newpar}
Given the observation graph \(\mathcal{G}_{\mathrm{obs}}\), the localization
task is to infer the damage coordinate from the measured pattern of path-wise
energy deviations. We denote this inverse mapping by
\begin{equation}
\hat{\mathbf{p}}^{(0)}
=
f\!\left(\mathcal{G}_{\mathrm{obs}}\right),
\label{eq:localization_problem}
\end{equation}
where \(f(\cdot)\)  denotes the initial
coordinate estimate estimate of the damage
location.

Inferring the damage location from this response pattern is, however,  weakly constrained when  sensing is sparse and damaged training samples are available
at only a limited set of spatial locations. Distinct defect locations may produce  similar
perturbation patterns  across the measured propagation paths, making the mapping from observed responses to damage coordinates ambiguous.
Consequently, accurate localization at damage locations included during
training does not guarantee that predictions at unseen locations are
consistent with the measured response across the sensing network.This ambiguity motivates an additional consistency requirement. A plausible
damage coordinate should not only be predicted by the inverse mapping but
should also reproduce the path-wise response pattern measured for the current
sample. To formalize this requirement, for a candidate damage location \(\mathbf{q}\in\Omega\), we consider the
forward  response mapping
\begin{equation}
g:\Omega\rightarrow[0,1]^{|\mathcal{P}|},
\qquad
\mathbf{q}\mapsto
\widehat{\Delta\mathbf{E}}(\mathbf{q}),
\label{eq:forward_response_relation}
\end{equation}
where the fixed transducer geometry is implicit. We define a candidate coordinate 
\(\mathbf{q}\) as response-consistent if
\begin{equation}
g(\mathbf{q})\approx\Delta\mathbf{E},
\label{eq:response_consistency_requirement}
\end{equation}
that is, if the path-wise energy-deviation pattern associated with the candidate
location agrees with the pattern measured across the sensing network for the current sample.

\end{newpar}

\section{Proposed Framework}
\label{sec:method}



Building on the graph representation and response-consistency formulation introduced above, \textit{WaveGraphNet} consists of two independently trained branches that model complementary mappings between damage location and the measured path-wise response. The inverse branch maps the observation graph
\(\mathcal{G}_{\mathrm{obs}}\) to an initial damage-coordinate estimate, whereas the forward branch maps a candidate damage location to the corresponding pattern of path-wise energy deviations.

At test time, the two branches are combined through coordinate refinement. Starting from the inverse prediction, the gradients obtained through the forward branch are used to update the candidate damage coordinate and reduce the discrepancy between the predicted and measured path-wise response patterns. 
In this way, the initial localization is refined according to the response-consistency criterion introduced in
Section~\ref{sec:localization_consistency}.

\begin{figure*}[!htpb]
    \centering
    \includegraphics[width=1.0\linewidth]{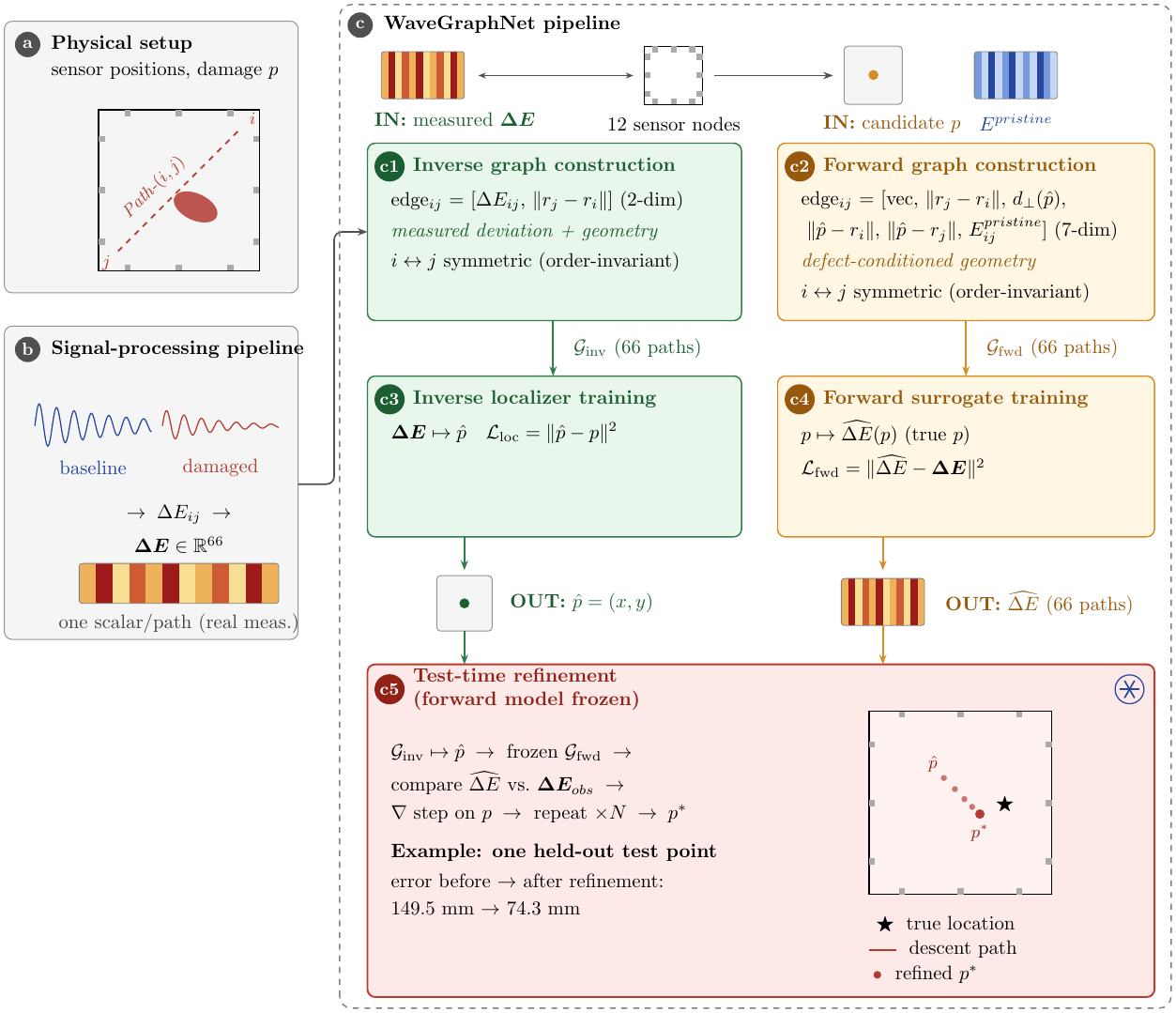}
    \caption{\new{\textbf{Overview of the proposed \textit{WaveGraphNet} framework.} \textbf{(a)} A CFRP plate instrumented with 12 boundary-mounted sensors. A defect at \(\mathbf{p}\) perturbs the guided-wave responses measured along the sensor-to-sensor propagation paths. \textbf{(b)} Comparison with the pristine responses gives one response-deviation value \(\Delta E_{ij}\) for each of the 66 measured paths. \textbf{(c1/c3)} The inverse branch maps the measured path-wise response pattern \(\Delta\mathbf{E}\) to an initial location \(\hat{\mathbf{p}}\). \textbf{(c2/c4)} The forward branch maps a candidate location \(\mathbf{q}\) to a predicted response pattern \(\widehat{\Delta\mathbf{E}}(\mathbf{q})\). \textbf{(c5)} During test-time refinement, the model parameters remain fixed and only the candidate coordinate is updated to reduce the difference between the predicted and measured response patterns.}}
    \label{fig:framework_overview}
\end{figure*}

\new{The two branches therefore learn complementary mappings between the path-wise response pattern and the damage location:
\[
\Delta\mathbf{E}\longrightarrow\hat{\mathbf{p}},
\qquad
\mathbf{q}\longrightarrow\widehat{\Delta\mathbf{E}}(\mathbf{q}).
\]
Rather than forming exact mathematical inverses, the two mappings serve distinct roles in the localization framework. The inverse branch directly estimates the damage location, while the forward branch learns the response pattern associated with a candidate location from data. The latter does not solve the governing guided-wave equations but provides a differentiable response-consistency criterion: a candidate location is considered more consistent with the measurement if its predicted path-wise response pattern closely matches the observed one. Figure~\ref{fig:framework_overview} summarizes the resulting framework.}

\subsection{Graph representation}
\label{sec:graph_representation}

\new{Both branches operate on the same graph topology, comprising the 12 transducer nodes and 66 measured propagation paths. Each node \(i\in V\) represents one transducer and is assigned its normalized in-plane spatial coordinate,}
\begin{equation}
\mathbf{x}_i=\mathbf{r}_i.
\end{equation}

\new{Each measured propagation path is represented in the graph by two directed edges, \((i,j)\) and \((j,i)\). Because only one measurement is available for each transducer pair, the underlying energy measurement, and hence the edge feature \(\Delta E_{ij}=\Delta E_{ji}\), is identical for both directed edges. This quantity serves as an edge feature for the inverse branch and the prediction target for the forward branch.}

\new{Using the same node and path sets, the inverse and forward graphs are defined as}
\begin{align}
\mathcal{G}_{\mathrm{inv}}
&=
\left(
V,\,
\mathcal{P},\,
\{\mathbf{x}_i\}_{i\in V},\,
\{\mathbf{e}^{\mathrm{inv}}_{ij}\}_{(i,j)\in\mathcal{P}}
\right),\\
\mathcal{G}_{\mathrm{fwd}}(\mathbf{q})
&=
\left(
V,\,
\mathcal{P},\,
\{\mathbf{x}_i\}_{i\in V},\,
\{\mathbf{e}^{\mathrm{fwd}}_{ij}(\mathbf{q})\}_{(i,j)\in\mathcal{P}}
\right).
\end{align}

\new{The two graphs differ only in their edge features. The inverse graph contains the measured response deviation and path length, whereas the forward graph contains the path geometry, the geometry of the candidate location, and the pristine response level of each path.}

\subsubsection{Inverse edge features}

\new{The inverse branch infers a defect location from the pattern of response changes across the sensing network. For each measured path \((i,j)\), \(\Delta E_{ij}\) is associated with the corresponding edge and quantifies its deviation from the pristine response. We additionally provide the path length which encodes the propagation distance between the two transducers and thereby supplies geometric information relevant to the measured response. The inverse edge feature is therefore}
\begin{equation}
\mathbf{e}^{\mathrm{inv}}_{ij}
=
\left[
\Delta E_{ij},\;
\|\mathbf{r}_j-\mathbf{r}_i\|_2
\right].
\label{eq:inverse_edge_features}
\end{equation}

\new{Because both directed edges of a path are built from the same underlying quantities, their edge features are identical by construction:}
\begin{equation}
\mathbf{e}^{\mathrm{inv}}_{ij}
=
\mathbf{e}^{\mathrm{inv}}_{ji}.
\label{eq:inverse_feature_symmetry}
\end{equation}

\new{This invariance is consistent with both the acquisition setting and the assumed reciprocal propagation regime. The dataset provides a single measured channel for each transducer pair, where changing the order of the transducers does not change the physical propagation path.} 

\subsubsection{Forward edge features}

\new{The forward branch performs the complementary mapping. Given a candidate damage location \(\mathbf{q}\), it predicts the response deviation associated with each measured path. 
The forward edge features are given by}
\begin{equation}
\begin{aligned}
\mathbf{e}^{\mathrm{fwd}}_{ij}(\mathbf{q})
=
\Big[
&\mathbf{r}_j-\mathbf{r}_i,\;
\|\mathbf{r}_j-\mathbf{r}_i\|_2,\;
d_{\perp}(\mathbf{q};i,j),\\
&\|\mathbf{q}-\mathbf{r}_i\|_2,\;
\|\mathbf{q}-\mathbf{r}_j\|_2,\;
E^{\mathrm{pristine}}_{ij}
\Big]
\in\mathbb{R}^{7}.
\end{aligned}
\label{eq:forward_edge_features}
\end{equation}

\new{The direction vector \(\mathbf{r}_j-\mathbf{r}_i\) encodes the orientation of the propagation path relative to the material axes, allowing the model to account for direction-dependent wave propagation in anisotropic CFRP. 
The path length $\|\mathbf{r}_j-\mathbf{r}_i\|_2$ encodes the propagation distance. The quantity \(d_{\perp}(\mathbf{q};i,j)\) denotes the shortest distance from the candidate damage location to the line segment joining the two transducers, while the two endpoint distances ($\|\mathbf{q}-\mathbf{r}_i\|_2$ and $\|\mathbf{q}-\mathbf{r}_j\|_2$) locate the candidate damage relative to the individual transducers. These quantities characterize the geometry of the candidate location relative to each propagation path.}
\new{The scalar \(E^{\mathrm{pristine}}_{ij}\) denotes the pristine (undamaged) response level of path \((i,j)\). The same value is assigned to the two directions \((i,j)\) and \((j,i)\), such that \(E^{\mathrm{pristine}}_{ij}=E^{\mathrm{pristine}}_{ji}\), because they correspond to a single measurement for the transducer pair \(i\) and \(j\).}

\new{Unlike the edge features in the inverse branch, the forward edge features are not invariant to interchanging nodes \(i\) and \(j\), because the direction vector changes sign and the two endpoint distances exchange positions. While, the prediction target remains unchanged, \(\Delta E_{ij}=\Delta E_{ji}\). The forward branch therefore enforces this invariance by construction at the encoding stage, using a DeepSets-style symmetric encoding~\citep{zaheer2017deep}, as described in Section~\ref{sec:forward_module}.}

\begin{table}[!h]
\centering
\caption{\new{Graph features used in \textit{WaveGraphNet}. Both branches use the same 12 nodes and 66 measured paths.}}
\label{tab:graph_features}
\begin{tabular}{p{0.18\linewidth} p{0.31\linewidth} p{0.41\linewidth}}
\toprule
\textbf{Graph component} & \textbf{Feature type} & \textbf{Definition} \\
\midrule
Node \(v_i\) & Node feature & \new{Sensor coordinate \(\mathbf{x}_i=\mathbf{r}_i\)} \\
\midrule
Inverse edge \(e^{\mathrm{inv}}_{ij}\) & \new{Measured and geometric features} & \new{Response deviation \(\Delta E_{ij}\) and path length \(\|\mathbf{r}_j-\mathbf{r}_i\|_2\)} \\
\midrule
Forward edge \(e^{\mathrm{fwd}}_{ij}\) & \new{Candidate-relative path features} & \new{Path direction, path length, candidate-to-path distance, candidate-to-sensor distances, and the pristine response of the path} \\
\bottomrule
\end{tabular}
\end{table}

\subsection{Inverse localization branch}
\label{sec:inverse_module}

\new{The inverse branch maps response pattern encoded by $\mathcal{G}_{\mathrm{inv}}$ to the initial damage-coordinate estimate:}
\begin{equation}
\hat{\mathbf{p}}^{(0)}
=
f_{\theta}(\mathcal{G}_{\mathrm{inv}}).
\label{eq:inverse_mapping}
\end{equation}

\new{Node and edge encoders  first map the transducer coordinates and path-wise features, respectively, into latent representations:}
\begin{align}
\mathbf{h}^{(0)}_i
&=
\phi_{\mathrm{node}}(\mathbf{r}_i),\\
\mathbf{h}^{(0)}_{ij}
&=
\phi_{\mathrm{enc}}
\left(
\Delta E_{ij},
\|\mathbf{r}_j-\mathbf{r}_i\|_2
\right).
\label{eq:edge_encoder}
\end{align}

\new{Because the inverse edge features are invariant to interchanging the two connected nodes, the initial edge embeddings inherit the same property:}
\begin{equation}
\mathbf{h}^{(0)}_{ij}
=
\mathbf{h}^{(0)}_{ji}.
\label{eq:initial_edge_symmetry}
\end{equation}

\new{To preserve this invariance to node interchange throughout message passing, each edge update depends on the two node embeddings only through their sum. At interaction layer \(\ell\), }
\begin{equation}
\mathbf{h}^{(\ell+1)}_{ij}
=
\mathbf{h}^{(\ell)}_{ij}
+
\psi^{(\ell)}_{\mathrm{edge}}
\left(
\mathbf{h}^{(\ell)}_i+\mathbf{h}^{(\ell)}_j,\,
\mathbf{h}^{(\ell)}_{ij}
\right).
\label{eq:edge_update}
\end{equation}

\new{Since summation is invariant to node interchange,}
\begin{equation}
\mathbf{h}^{(\ell)}_i+\mathbf{h}^{(\ell)}_j
=
\mathbf{h}^{(\ell)}_j+\mathbf{h}^{(\ell)}_i,
\label{eq:endpoint_exchange_symmetry}
\end{equation}
\new{the resulting edge update is identical under $i \leftrightarrow j$. Node-order invariance is therefore enforced by construction rather than learned from duplicated representations of both path orientations. This node interchange invariance constitutes the physics-consistent symmetry imposed in the inverse branch.}

\new{The updated edge representations are subsequently aggregated at each transducer node:}
\begin{equation}
\mathbf{h}^{(\ell+1)}_i
=
\mathbf{h}^{(\ell)}_i
+
\psi^{(\ell)}_{\mathrm{node}}
\left(
\mathbf{h}^{(\ell)}_i,\,
\sum_{j\in\mathcal{N}(i)}
\mathbf{h}^{(\ell+1)}_{ij}
\right).
\label{eq:node_update}
\end{equation}

\new{The aggregation over \(\mathcal{N}(i)\) integrates information from all measured propagation paths incident to transducer \(i\). Repeating the edge and node updates over \(L_{\mathrm{inv}}=3\) interaction layers progressively propagates and combines path-wise response information across the sensing network.}

\subsubsection{Path-constrained coordinate decoder}

\new{A single propagation path cannot uniquely identify a two-dimensional defect location but provides partial spatial evidence along the line joining its two sensors. The decoder therefore allows each path to propose a candidate location constrained to this segment and subsequently aggregates the path-level proposals into a single damage-coordinate estimate.}

\new{After the \(L_{\mathrm{inv}}=3\) \footnote{Minimum number of message passing steps required for information to pass between edges that do not share a transducer.} interaction layers, the final node embedding \(\mathbf{h}^{(L_{\mathrm{inv}})}_i\) summarizes the information propagated to transducer \(i\), while \(\mathbf{h}^{(L_{\mathrm{inv}})}_{ij}\) encodes path \((i,j)\) in the context of the full sensing network. The decoder uses these final embeddings to to generate a path-level coordinate proposal. For path \((i,j)\), a shared multilayer perceptron \(g(\cdot)\) assigns a scalar score to each node based on the node and shared path embeddings. A larger score places the proposed point closer to that node. The same function is applied to both nodes, preventing a fixed preference for sensor \(i\) or sensor \(j\). A softmax converts the two scores into non-negative weights \(w_i\) and \(w_j\), with \(w_i+w_j=1\):}

\begin{equation}
\begin{aligned}
(w_i,w_j)
&=
\operatorname{softmax}
\Big(
g(\mathbf{h}^{(L_{\mathrm{inv}})}_i,
  \mathbf{h}^{(L_{\mathrm{inv}})}_{ij}),\\
&\hspace{26mm}
g(\mathbf{h}^{(L_{\mathrm{inv}})}_j,
  \mathbf{h}^{(L_{\mathrm{inv}})}_{ij})
\Big),\\
\mathbf{q}_{ij}
&=
w_i\mathbf{r}_i+w_j\mathbf{r}_j.
\end{aligned}
\label{eq:endpoint_vote}
\end{equation}

\new{Because \(w_i,w_j\geq 0\) and \(w_i+w_j=1\), the path-level proposal \(\mathbf{q}_{ij}\) is a convex combination of the two transducer coordinates and therefore lies on the segment joining transducers \(i\) and \(j\). Applying the same scoring function to both nodes also preserves the invariance to node interchange: interchanging \(i\) and \(j\) leaves \(\mathbf{q}_{ij}\) unchanged.}

\new{
The 66 path-level damage coordinate proposals is then be combined into one coordinate. To do this, a shared learnable multilayer perceptron \(\phi_{\mathrm{w}}(\cdot)\) maps the final embedding of each path to a scalar importance score \(a_{ij}\). A softmax applied across all measured paths converts these scores into non-negative path weights \(W_{ij}\) that sum to one. The final damaged-location estimate is the weighted average of the path-level proposals:}
\begin{equation}
\begin{aligned}
a_{ij}
&=
\phi_{\mathrm{w}}
\left(
\mathbf{h}^{(L_{\mathrm{inv}})}_{ij}
\right),\\
W_{ij}
&=
\frac{\exp(a_{ij})}
{\displaystyle
\sum_{(k,l)\in\mathcal{P}}
\exp(a_{kl})
},\\
\mathbf{p}_{\mathrm{conv}}
&=
\sum_{(i,j)\in\mathcal{P}}
W_{ij}\mathbf{q}_{ij}.
\end{aligned}
\label{eq:edge_pooling}
\end{equation}

\new{The weight \(W_{ij}\) determines the contribution of path \((i,j)\) to the final estimate. Paths assigned larger importance scores contribute more strongly, while paths assigned smaller scores have less influence. Because \(W_{ij}\geq0\) and \(\sum_{(i,j)\in\mathcal{P}}W_{ij}=1\), \(\mathbf{p}_{\mathrm{conv}}\) is a convex combination of the 66 path-level proposals.}

\new{The model can therefore assign larger weights to paths that provide stronger localization evidence. Since the path weights are non-negative and sum to one, \(\mathbf{p}_{\mathrm{conv}}\) lies within the convex hull of the sensor positions. This constrains damaged predictions to the instrumented region but also means that the inverse decoder cannot represent a damaged location outside that convex hull.}

\new{Because \(\mathbf{p}_{\mathrm{conv}}\) is confined to the sensor convex hull, it cannot represent the no-damage coordinate \(\mathbf{p}_{\mathrm{ud}}\) located outside the plate. We therefore introduce a soft gate \(d\in[0,1]\), computed from the average of the final node embeddings, to combine \(\mathbf{p}_{\mathrm{conv}}\) with \(\mathbf{p}_{\mathrm{ud}}\):}

\begin{equation}
d
=
\sigma
\left(
\phi_{\mathrm{dmg}}
\left(
\frac{1}{N}
\sum_{i=1}^{N}
\mathbf{h}^{(L_{\mathrm{inv}})}_i
\right)
\right).
\label{eq:damage_gate}
\end{equation}

\new{This value interpolates between the damaged-location estimate and the fixed no-damage coordinate:}
\begin{equation}
\hat{\mathbf{p}}^{(0)}
=
d\,\mathbf{p}_{\mathrm{conv}}
+
(1-d)\mathbf{p}_{\mathrm{ud}}.
\label{eq:coordinate_decoder}
\end{equation}

\new{For damaged samples, the localization target lies inside the plate and encourages \(d\) to approach one. For undamaged samples, the target is \(\mathbf{p}_{\mathrm{ud}}\) and encourages \(d\) to approach zero.}

\subsection{\new{Forward response model}}
\label{sec:forward_module}

\new{The inverse branch predicts a damage location directly from the measured response pattern, but  does not assess whether the predicted location can reproduce the measured response pattern. The forward branch provides this complementary response-consistency assessment. Given a candidate coordinate \(\mathbf{q}\), it predicts an energy-deviation value for each measured propagation path. Because the two directed edges \((i,j)\) and \((j,i)\) associated with a path correspond to the same physical measurement, the prediction on both directed edges can be compared directly with the measured energy deviation, \(\Delta E_{ij}\), on that path.}
\begin{equation}
\widehat{\Delta\mathbf{E}}(\mathbf{q})
=
g_{\psi}
\left(
\mathcal{G}_{\mathrm{fwd}}(\mathbf{q})
\right).
\label{eq:forward_mapping}
\end{equation}


\new{The forward branch computes \(g_{\psi}\) using the same encode--process--decode structure as the inverse branch, applied instead to the forward graph. Unlike the inverse edge features, however, the raw forward edge features \(\mathbf{e}^{\mathrm{fwd}}_{ij}(\mathbf{q})\) are not invariant to interchanging the connected nodes (Section~\ref{sec:graph_representation}), because they contain directional geometric information. To obtain an order-invariant representation while retaining this information, each measured path is encoded in both directions using the same shared encoder, and the resulting embeddings are averaged:}

\begin{equation}
\mathbf{u}^{(0)}_{ij}
=
\mathbf{u}^{(0)}_{ji}
=
\frac{1}{2}
\left(
\phi_{\mathrm{fwd,enc}}\!\left(\mathbf{e}^{\mathrm{fwd}}_{ij}(\mathbf{q})\right)
+
\phi_{\mathrm{fwd,enc}}\!\left(\mathbf{e}^{\mathrm{fwd}}_{ji}(\mathbf{q})\right)
\right).
\label{eq:forward_edge_encoder}
\end{equation}
\new{Because averaging is invariant to the ordering of its inputs, the resulting initial edge embedding is invariant to node interchange by construction, even though the underlying directional feature vectors are not.}

\new{The forward model preserves this node-interchange invariance during message passing using the same symmetric node aggregation as the inverse model:}
\begin{equation}
\mathbf{u}^{(\ell+1)}_{ij}
=
\mathbf{u}^{(\ell)}_{ij}
+
\psi^{(\ell)}_{\mathrm{fwd,edge}}
\left(
\mathbf{u}^{(\ell)}_i+\mathbf{u}^{(\ell)}_j,\,
\mathbf{u}^{(\ell)}_{ij}
\right).
\label{eq:forward_edge_update}
\end{equation}

\new{Node aggregation then combines the updated edge embeddings from all incident edges at each node:}
\begin{equation}
\mathbf{u}^{(\ell+1)}_i
=
\mathbf{u}^{(\ell)}_i
+
\psi^{(\ell)}_{\mathrm{fwd,node}}
\left(
\mathbf{u}^{(\ell)}_i,\,
\sum_{j\in\mathcal{N}(i)}
\mathbf{u}^{(\ell+1)}_{ij}
\right).
\label{eq:forward_node_update}
\end{equation}

\new{In Eq.~\eqref{eq:forward_edge_update}, the two node embeddings enter the edge update only through their sum, which is unchanged under node interchange because  addition is commutative. Starting from the symmetric initial embedding (Eq.~\eqref{eq:forward_edge_encoder}), applying this  update at each layer keeps the edge embedding symmetric throughout all \(L_{\mathrm{fwd}}\) layers. The forward branch uses the same number of interaction layers as the inverse branch, \(L_{\mathrm{fwd}}=L_{\mathrm{inv}}=3\), since both operate over the same complete sensing graph.}

\new{After the interaction layers, an edge decoder predicts the energy deviation of each edge. Since \(\mathbf{u}^{(L_{\mathrm{fwd}})}_{ij}=\mathbf{u}^{(L_{\mathrm{fwd}})}_{ji}\), the decoder gives the same prediction in either direction, which is compared against the single measured value \(\Delta E_{ij}\) for that path:}
\begin{equation}
\widehat{\Delta E}_{ij}(\mathbf{q})
=
\phi_{\mathrm{fwd,dec}}
\left(
\mathbf{u}^{(L_{\mathrm{fwd}})}_{ij}
\right).
\label{eq:forward_edge_prediction}
\end{equation}

\new{Concatenating predictions for all 66 paths gives}
\begin{equation}
\widehat{\Delta\mathbf{E}}(\mathbf{q})
=
\left[
\widehat{\Delta E}_{ij}(\mathbf{q})
\right]_{(i,j)\in\mathcal{P}}.
\label{eq:forward_vector_prediction}
\end{equation}


\subsection{Training}
\label{sec:training_strategy}

\new{The inverse and forward branches are trained separately, each with its own objective and its own subset of training samples; neither branch's loss depends on the other. The two are only combined afterward, at inference time, through the test-time refinement procedure described in Section~\ref{sec:test_time_refinement}.}

\paragraph{Inverse branch}

\new{The inverse branch is trained on both damaged and undamaged samples, using the mean squared coordinate-localization error below as its objective. The target \(\mathbf{p}_{\mathrm{tar}}\) is the true defect coordinate for a damaged sample, and the fixed no-damage coordinate \(\mathbf{p}_{\mathrm{ud}}\), located outside the plate, for an undamaged sample:}
\begin{equation}
\mathcal{L}_{\mathrm{loc}}
=
\left\|
\hat{\mathbf{p}}^{(0)}
-
\mathbf{p}_{\mathrm{tar}}
\right\|_2^2.
\label{eq:loc_loss}
\end{equation}

\paragraph{Forward branch}

\new{The forward branch is trained separately using damaged samples only. For each training sample, the true defect coordinate \(\mathbf{p}^{(n)}\) is used to construct the forward graph. Both \(E^{\mathrm{pristine}}_{ij}\) and the measured response deviation \(\Delta E_{ij}\) are divided by a global scaling constant \(s\) -- the mean pristine energy level over all paths in the training data -- before being used as model input and regression target, respectively. The predicted path-wise response is compared directly with this scaled measured response:}
\begin{equation}
\mathcal{L}_{\mathrm{fwd}}
=
\frac{1}{|\mathcal{D}_{\mathrm{tr}}||\mathcal{P}|}
\sum_{n\in\mathcal{D}_{\mathrm{tr}}}
\sum_{(i,j)\in\mathcal{P}}
\left(
\widehat{\Delta E}_{ij}
\left(
\mathbf{p}^{(n)}
\right)
-
\frac{\Delta E_{ij}^{(n)}}{s}
\right)^2,
\label{eq:fwd_loss}
\end{equation}
\new{where \(\mathcal{D}_{\mathrm{tr}}\) denotes the damaged training samples. Undamaged samples are excluded because they have no physical defect coordinate from which to construct the candidate-relative forward features.}

\paragraph{Model selection}

\new{The two checkpoints are selected independently. The inverse checkpoint is selected using localization error on the validation partition, while the forward checkpoint is selected using path-wise response-prediction error on damaged validation samples. The spatially held-out test regions are not used for checkpoint selection.}

\subsection{Inference pipeline}
\label{sec:inference_pipeline}

\new{The standalone configuration requires one pass through the inverse branch:}
\begin{equation}
\hat{\mathbf{p}}^{(0)}
=
f_{\theta}
\left(
\mathcal{G}_{\mathrm{inv}}
\right).
\label{eq:standalone_inference}
\end{equation}

\new{This initial prediction is reported as \textit{WaveGraphNet, standalone}.}

\subsection{Test-time refinement}
\label{sec:test_time_refinement}

\new{During test time, \(\hat{\mathbf{p}}^{(0)}\) from the trained inverse model provides the starting location. The forward model predicts the 66 path responses associated with this location and compares them with the measured responses. If the two patterns differ, the candidate coordinate is moved in the direction that reduces the mismatch.}

\new{For a candidate coordinate \(\mathbf{q}\), the response mismatch is}
\begin{equation}
\mathcal{J}(\mathbf{q})
=
\frac{1}{|\mathcal{P}|}
\left\|
g_{\psi}
\left(
\mathcal{G}_{\mathrm{fwd}}(\mathbf{q})
\right)
-
\Delta\mathbf{E}
\right\|_2^2.
\label{eq:refinement_objective}
\end{equation}

\new{This objective gives a direct consistency test. A smaller value means that the candidate location produces a path-wise response pattern closer to the measured pattern. Starting from \(\mathbf{q}^{(0)}=\hat{\mathbf{p}}^{(0)}\), Adam optimizer updates the coordinate according to}
\begin{equation}
\mathbf{q}^{(n)}
=
\operatorname{AdamUpdate}_{\eta_r}
\left(
\mathbf{q}^{(n-1)},
\nabla_{\mathbf{q}}
\mathcal{J}
\left(
\mathbf{q}^{(n-1)}
\right)
\right),
\qquad
n=1,\ldots,N_r.
\label{eq:test_time_refinement}
\end{equation}

\new{The reported experiments use \(N_r=60\) refinement steps and a learning rate of \(\eta_r=0.01\). These values are empirical optimization settings. Gradients pass through the forward model to the candidate coordinate.}

\new{Because the learned response objective is non-convex, the objective value does not necessarily decrease at every iteration. The method therefore returns the candidate with the smallest response mismatch rather than automatically returning the final candidate:}
\begin{equation}
n^\star
=
\arg\min_{n\in\{0,\ldots,N_r\}}
\mathcal{J}
\left(
\mathbf{q}^{(n)}
\right),
\qquad
\mathbf{p}^{\star}
=
\mathbf{q}^{(n^\star)}.
\label{eq:best_refined_coordinate}
\end{equation}

\new{The resulting coordinate is reported as \textit{WaveGraphNet, + refinement}. Unlike the standalone configuration, this variant requires the forward model at deployment and performs 60 additional forward and backward evaluations for each refined sample. In the reported localization experiments, refinement is applied only to damaged samples. Undamaged samples are evaluated using the standalone inverse prediction when computing the false-positive rate; consequently, refinement does not act as a separate damage-detection procedure.}

\section{Case study: Guided-wave localization in CFRP plates}
\label{sec:case_study}

The proposed framework is evaluated on the openly available Open Guided Waves benchmark dataset, referred to here as OGW-1, for guided-wave structural health monitoring on a composite plate \cite{ogw1}. This dataset was collected and published independently of the present study, and full experimental details are provided in the original benchmark paper \cite{ogw1}. OGW-1 is an openly available guided-wave benchmark platform and was not  specifically designed for the present study. For the present evaluation, we define \new{three} spatial hold-out splits on this dataset with increasing extrapolation difficulty. \new{These splits} serve as controlled experimental testbeds for assessing whether  the proposed inductive bias improves localization performance in unseen damage regions under limited spatial training coverage, rather than for demonstrating broad statistical coverage of operating conditions.

\subsection{Problem setup}
\label{sec:case_problem_setup}

The specimen is a quasi-isotropic CFRP plate representative of lightweight aerospace structures. It is manufactured from \textit{Hexply\textsuperscript{\textregistered} M21/34\%/UD134/ T700/300} prepreg material, has nominal dimensions of $500\times500~\mathrm{mm}^2$ and a thickness of $2~\mathrm{mm}$. The laminate consists of 16 plies arranged in a symmetric and balanced stacking sequence $[\ang{45}/0 \ang{-45}/90]_s$, providing quasi-isotropic in-plane stiffness through equal representation of $0^\circ$, $\pm45^\circ$, and $90^\circ$ fiber orientations. The plate is instrumented with 12 co-bonded DuraAct piezoelectric transducers arranged in two rows along the plate boundaries, yielding a sparse edge-mounted sensing layout typical of large-area monitoring applications \cite{capineri2021guided_wave_sensors_review}. Each transducer has a diameter of $5~\mathrm{mm}$ and a thickness of $0.2~\mathrm{mm}$.

Measurements are acquired in pitch--catch mode, in which one transducer excites the plate and another records the received response. As a result, localization must be inferred from path-wise perturbations across the sensor network rather than from full-field wavefield measurements. To enable repeatable damage scenarios without permanently modifying the specimen, the benchmark uses a reversible pseudo-defect consisting of an aluminum disk attached with tacky tape. This surrogate does not reproduce every aspect of real in-service damage, but it does create repeatable local scattering, attenuation, and travel-time perturbations suitable for benchmarking localization algorithms under controlled conditions.

\begin{figure*}[!h]
    \centering
    \includegraphics[width=\linewidth]{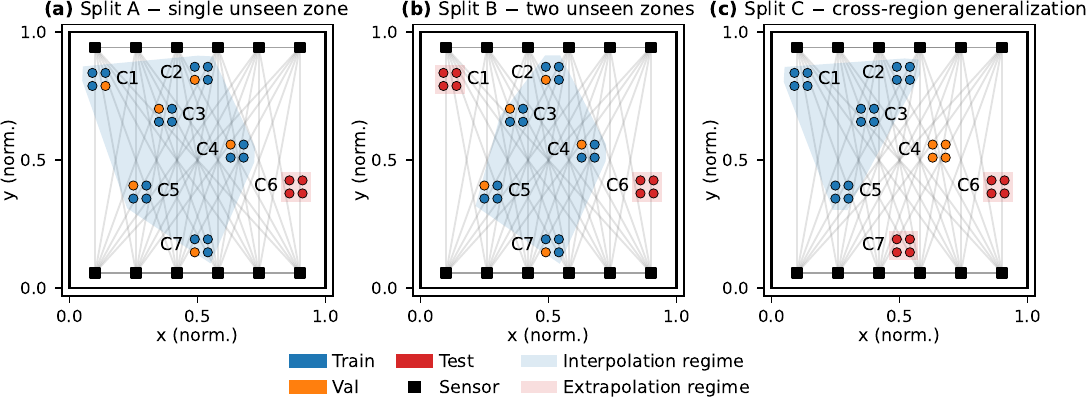}
    \caption{\new{\textbf{Spatial train--validation--test splits on the OGW-1 SHM
    Plate, in increasing order of extrapolation difficulty.} The 28 defect
    locations form 7 tight spatial clusters (C1--C7); The shaded blue
    region marks the convex hull of the training damage locations (the
    \emph{interpolation regime}); shaded red regions mark held-out test
    clusters lying outside this hull (the \emph{extrapolation regime}).
    \textbf{(a)} Split~A: a single held-out test cluster (C6,
    D$_{21}$--D$_{24}$); training and validation samples are drawn from
    the remaining six clusters, spread across nearly the entire plate.
    \textbf{(b)} Split~B: two held-out test clusters (C1, D$_1$--D$_4$,
    and C6, D$_{21}$--D$_{24}$), both outside the training hull --
    a harder, two-region extrapolation setting.
    \textbf{(c)} Split~C: the hardest setting. Training is confined to a
    single contiguous region (C1, C2, C3, C5); validation uses a cluster
    (C4) that is spatially separate from training rather than interleaved
    with it, so checkpoint selection itself already requires
    generalization; test comprises two further clusters (C6, C7) beyond
    both training and validation. Unlike Splits~A and~B, no baseline model
    suite was evaluated on Split~C (see Section~\ref{sec:results}); it is
    used to stress-test the proposed method and the training-free
    baseline described in Section~\ref{sec:implementation_details}.}}
    \label{fig:shm_setup_case}
\end{figure*}

\subsection{Dataset construction and preprocessing}
\label{sec:case_dataset_construction}

For each experiment, one transducer acts as an actuator, and responses are recorded for the remaining channels, producing the path-wise signals $s_{ij}(t)$ introduced in Section~\ref{sec:problem_formulation}. As described in the OGW-1 acquisition protocol \cite{ogw1}, each unordered transducer pair is measured only once; consequently, $(i,j)$ and $(j,i)$ are not considered distinct observations. For a system with 12 transducers, this results in 66 unique propagation paths per sample. Accordingly, each sample is represented on a fixed graph topology, with fixed node coordinates and a consistent set of path-wise measurements throughout the study.

The campaign comprises repeated pristine acquisitions as well as measurements obtained by placing a pseudo-defect at multiple spatial locations. In line with the scope of this study, only single-defect cases are considered. Multi-defect examples available in the repository are excluded, as the learning target is a single  coordinate $\mathbf{p}\in\Omega$ per sample. After this filtering, the resulting dataset consists of 28 damaged samples and 60 pristine samples. While these sample sizes are small by conventional machine-learning standards, they are appropriate for the present study, which is to compare modeling choices under controlled data-scarcity conditions rather than to demonstrate saturated benchmark performance.

Prior to feature extraction, the raw voltage traces are high-pass filtered using a third-order Butterworth filter with a cutoff frequency of $20~\mathrm{kHz}$ in order to suppress low-frequency drift and background components while preserving the guided-wave content around the selected $100~\mathrm{kHz}$ excitation band. Each retained sample is  initially represented as a $13{,}108\times66$ matrix of time-domain guided-wave measurements, where 13,108 denotes the number of temporal samples per propagation path and 66 corresponds to the number of unique paths induced by the 12-transducer configuration. A path-wise pristine reference signal, $\bar{s}_{ij}(t)$, is estimated from the healthy measurements and subtracted from each sample to obtain the differential response defined in Eq.~\eqref{eq:differential_signal}. The same reference construction procedure is applied consistently across all samples, ensuring that damaged measurements are compared against a common healthy baseline. \new{The resulting differential signals are subsequently transformed into the frequency domain and summarized into the path-wise energy-deviation scalar $\Delta E_{ij}$ (Eq.~\eqref{eq:deltaE}).} Amplitude normalization is performed using statistics computed exclusively from the training data without incorporating information from the held-out defect regions during the normalization process.

\new{For the forward branch, the observed consistency target $\Delta \mathbf{E}$ is computed from the same damaged and pristine spectral amplitudes using Eq.~\eqref{eq:deltaE}, over the same full measured path set $\mathcal{P}$ used by the inverse branch (Section~\ref{sec:graph_representation_one_meas}). The two branches therefore share the same transducer layout and the same path connectivity, differing only in their edge feature definitions (Section~\ref{sec:graph_representation}) and in how that shared quantity is used: as the inverse branch's input, and as the forward branch's supervision target.}



\subsection{Train--test protocol and evaluation setting}
\label{sec:case_train_test}

\begin{newpar}
\new{The method is evaluated using the three spatial hold-out splits shown in Fig.~\ref{fig:shm_setup_case}. In each split, the test locations form clusters that are fully isolated from training and validation and geometrically separated from the training locations. All repeated measurements from the same damage location remain in the same partition.}

In Split~A, cluster D$_{21}$--D$_{24}$ is reserved for testing. From each of the remaining six clusters, one location is used for validation and the other three are used for training. Split~B reserves two spatially separated corner clusters, D$_1$--D$_4$ and D$_{21}$--D$_{24}$, for testing and applies the same train--validation assignment to the remaining five clusters. Split~C imposes a stronger spatial separation: training uses D$_1$--D$_{12}$ and D$_{17}$--D$_{20}$, validation uses D$_{13}$--D$_{16}$, and testing uses D$_{21}$--D$_{28}$. Thus, checkpoint selection in Split~C also requires generalization beyond the training region.

The spatial hold-out protocol is more demanding than a random sample-wise split because the test coordinates lie in regions not represented during training. It therefore better reflects the deployment gap in SHM, where performance measured by interpolation among nearby damage locations can overestimate generalization to previously unobserved regions \cite{cawley2018closing_gap}.

Damaged training samples are combined with the undamaged samples assigned to the training partition. Undamaged samples use the external target \(\mathbf{p}_{\mathrm{ud}}=(-0.5,-0.5)\), allowing damaged and undamaged cases to be handled through the same coordinate-regression objective. Normalization statistics are computed from the training partition only, and all models use the same splits and preprocessing. During testing, held-out damaged samples are used to evaluate localization, while undamaged samples are used to evaluate whether predictions remain outside the admissible plate domain.

For held-out damaged samples, localization error is computed between the predicted coordinate \(\hat{\mathbf{p}}\) and the true defect coordinate \(\mathbf{p}\). Errors are reported in normalized coordinates and in millimeters. Results are presented separately for locations represented by the training coverage and for the unseen regions defined by Splits~A, B, and C. Because the benchmark contains only a limited number of distinct test locations, the results are interpreted primarily as relative comparisons among models under the same evaluation protocol rather than as comprehensive estimates of field performance.
\end{newpar}

\section{Implementation details and model configurations}
\label{sec:implementation_details}

\begin{table}[!h]
\centering
\caption{Architecture summary for all compared models. $d$: hidden
dimension; \emph{Pool}: graph-level aggregation operator;
BiLSTM: bidirectional long short-term memory;
MP: message-passing layer; GAT: graph attention network layer;
LayerNorm: layer normalization; ELU: exponential linear unit.}
\label{tab:architecture_summary}
\footnotesize
\setlength{\tabcolsep}{4pt}
\renewcommand{\arraystretch}{1.15}
\begin{tabularx}{\linewidth}{@{}
  >{\raggedright\arraybackslash}p{0.14\linewidth}
  >{\raggedright\arraybackslash}p{0.24\linewidth}
  >{\raggedright\arraybackslash}p{0.22\linewidth}
  >{\centering\arraybackslash}p{0.05\linewidth}
  >{\raggedright\arraybackslash}p{0.16\linewidth}
  >{\centering\arraybackslash}p{0.09\linewidth}
@{}}
\toprule
\textbf{Model} & \textbf{Encoder} & \textbf{Processor} &
$\mathbf{d}$ & \textbf{Layers $\times$ Heads} & \textbf{Pool} \\
\midrule
1D-CNN
  & 5-block Conv1d ($16{\to}256$, kernel~3, ReLU, MaxPool)
  & —
  & 256 & 5 conv & AvgPool \\
\addlinespace
LSTM
  & —
  & 3-layer BiLSTM + self-attention
  & 256 & 3 BiLSTM & Self-attn \\
\addlinespace
GNN-MLP
  & 4-layer MLP (static edge)
  & MP, mean aggregation
  & 256 & 4 MP & Mean \\
\addlinespace
GAT
  & 4-layer MLP (static edge)
  & GAT
  & 256 & 4 GAT $\times$ 16 & Attn. \\
\midrule
\textit{WGN} (inv.)
  & \new{MLP on $[\Delta E_{ij}, \|\mathbf{r}_j{-}\mathbf{r}_i\|_2]$ (2-dim, order-invariant)}
  & \new{3$\times$ order-invariant MP (sum of node embeddings)}
  & 256 & \new{3 MP} & \new{Conv. vote} \\
\addlinespace
\textit{WGN} (fwd.)
  & MLP on \new{7-dim} geometric \new{$+$ pristine-energy} edge features
  & \new{3$\times$} MP with residual connections
  & \new{256} & \new{3 MP} & Edge \\
\bottomrule
\end{tabularx}
\smallskip
\noindent\footnotesize
1D-CNN, LSTM, GNN-MLP, and GAT use a 3-layer MLP regression head predicting the 2D defect
coordinate. \textit{WGN}~(fwd.) outputs
$\widehat{\Delta\mathbf{E}}\in\mathbb{R}^{66}$ via a 2-layer MLP. \new{\textit{WGN}~(inv.) instead decodes via the two-stage, geometry-constrained convex-combination scheme of Eqs.~\eqref{eq:endpoint_vote}--\eqref{eq:coordinate_decoder}, gated by a learned damage probability against the no-damage sentinel $\mathbf{p}_{\mathrm{ud}}$ -- not a pooled regression head.} Dropout: LSTM~0.3, \textit{WGN}~(inv.)~0.2,
all others~0.
\end{table}

\begin{table}[!htbp]
\centering
\caption{Shared training settings and model-specific hyperparameters used in the experiments.}
\label{tab:training_protocol}
\footnotesize
\setlength{\tabcolsep}{5pt}
\renewcommand{\arraystretch}{1.15}
\begin{tabularx}{\linewidth}{@{}
  >{\raggedright\arraybackslash}p{0.34\linewidth}
  >{\raggedright\arraybackslash}X
@{}}
\toprule
\textbf{Setting} & \textbf{Value} \\
\midrule
\multicolumn{2}{@{}l}{\textit{Shared across all models}} \\
\addlinespace
Optimizer        & Adam \\
LR schedule      & ReduceLROnPlateau (factor~0.8, patience~20) \\
Early stopping   & \new{val-checked every 2 epochs; stop after 50 evaluations (100 epochs) without improvement} \\
Seeds            & $\{0,\,1,\,42\}$ \\
Frequency band   & 256 bins, $69.4$--$128\,\text{kHz}$ (FFT) \\
\midrule
\multicolumn{2}{@{}l}{\new{\textit{Per-model batch size / learning rate (locked via grid search, Section~\ref{sec:implementation_details})}}} \\
\addlinespace
\new{1D-CNN, LSTM, GNN-MLP, GAT} & \new{batch size 8, learning rate $10^{-3}$ (raw and $\Delta E$ input alike)} \\
\new{\textit{WGN} inverse branch} & \new{batch size~64, $\eta=10^{-4}$, loss: $\mathcal{L}_{\mathrm{loc}}$ (Eq.~\eqref{eq:loc_loss})} \\
\new{\textit{WGN} forward branch} & \new{batch size~128, $\eta=10^{-4}$, loss: $\mathcal{L}_{\mathrm{fwd}}$ (Eq.~\eqref{eq:fwd_loss})} \\
\new{Test-time refinement} & \new{$N=60$ Adam steps, $\eta_r=0.01$ (Eq.~\eqref{eq:test_time_refinement}); applied only at inference} \\
\bottomrule
\end{tabularx}
\end{table}

This section describes the experimental implementation and comparative evaluation setup used to assess the proposed framework. The localization problem formulation, graph construction procedure, coupled inverse--forward architecture, and training objectives were introduced in Sections~\ref{sec:problem_formulation} and \ref{sec:method}, while Sections~\ref{sec:case_study}--\ref{sec:case_train_test} detailed the dataset, preprocessing pipeline, spatial hold-out protocol, and evaluation metrics. Here, we focus specifically on the implementation choices required to reproduce the experiments and on the configurations of the compared models.

All models are trained and evaluated using the same preprocessing pipeline, frequency-band selection, normalization procedure, spatial split definitions, and random seeds. This ensures that observed performance differences arise from the model architecture and training objectives rather than from variations in data preparation. \new{The comparison includes two non-graph baselines, two graph-based baselines, \textit{WaveGraphNet} standalone (the independently trained inverse branch alone), and \textit{WaveGraphNet} with test-time refinement (the same standalone prediction, additionally corrected at inference by the independently trained, forward branch), together with RAPID, a classical training-free imaging baseline described in Section~\ref{sec:rapid_baseline}.}

The non-graph baselines, namely the one-dimensional convolutional neural network (1D-CNN) \cite{wang2017time} and long short-term memory network (LSTM) \cite{hochreiter1997long}, operate on ordered path-wise inputs without explicitly encoding the transducer-network geometry. In contrast, the graph-based baselines, namely the graph neural network with multilayer-perceptron edge encoder (GNN-MLP) \cite{battaglia2018relational_inductive_biases} and the graph attention network (GAT) \cite{velickovic2018gat}, use the same inverse graph representation as \textit{WaveGraphNet} but are trained solely with the coordinate localization objective. \new{All four are standard, widely used architectures from the general time-series and graph-learning literature, so that any advantage observed for \textit{WaveGraphNet} cannot be attributed to a weak or non-standard baseline.}

\begin{newpar}
Each of the four learned baselines is evaluated using two signal representations. In the \emph{raw-spectral} condition, each path is represented by its normalized amplitude \(\tilde{A}_{ij}(f_k)\) and phase \(\phi_{ij}(f_k)=\angle\mathcal{F}\{\delta s_{ij}\}(f_k)\) at the \(K\) selected frequency bins, giving \(2K\) values per path. In the \(\Delta E\) condition, these values are replaced by the single response-deviation index \(\Delta E_{ij}\) defined in Eq.~\eqref{eq:deltaE}. All other model settings remain unchanged. Comparing the two conditions shows the effect of the signal representation, while comparing the models under the same \(\Delta E\) input shows the effect of the architecture.
\end{newpar}

\new{To isolate the effect of test-time refinement, we evaluate \textit{WaveGraphNet} in two configurations: standalone, using only the independently trained inverse branch (Eq.~\eqref{eq:inverse_mapping}), and with refinement, in which the independently trained forward branch additionally corrects the standalone prediction at inference time (Section~\ref{sec:test_time_refinement}). The two branches are never trained jointly (Section~\ref{sec:training_strategy}); refinement is applied purely at test time, on top of an otherwise unchanged standalone model.}

\begin{newpar}
Hyperparameters for the four learned baselines were selected using a common grid search. The candidate batch sizes were \(8\), \(16\), \(32\), and \(64\), and the candidate learning rates were \(10^{-4}\), \(3\times10^{-4}\), and \(10^{-3}\). For the LSTM, the batch size was restricted to \(8\) because larger batches exceeded the available GPU memory. Validation MAE was aggregated across the four baselines, and one shared configuration was selected to provide the same tuning budget for every model. This procedure selected a batch size of \(8\) and a learning rate of \(10^{-3}\), which were used for both the raw-spectral and \(\Delta E\) input conditions.

The hyperparameters of the inverse and forward \textit{WaveGraphNet} branches were selected using an analogous validation-based sweep and a comparable search budget. Table~\ref{tab:architecture_summary} summarizes the model architectures, while Table~\ref{tab:training_protocol} reports the selected batch sizes, learning rates, and shared optimization settings.
\end{newpar}

\subsection{RAPID: a training-free imaging baseline}
\label{sec:rapid_baseline}

\begin{newpar}
RAPID (Reconstruction Algorithm for Probabilistic Inspection of Damage) is a classical guided-wave imaging method introduced by Van Velsor et al.~\cite{song2025unsupervised}. Its basic idea is simple: if the response of a sensor path changes, damage is more likely to be located near that path. RAPID applies this idea to every measured path and adds their contributions to form an image of the plate. It does not train a localization model.

\begin{newpar}
\begin{figure*}[h!]
    \centering
    \includegraphics[width=1.0\textwidth]{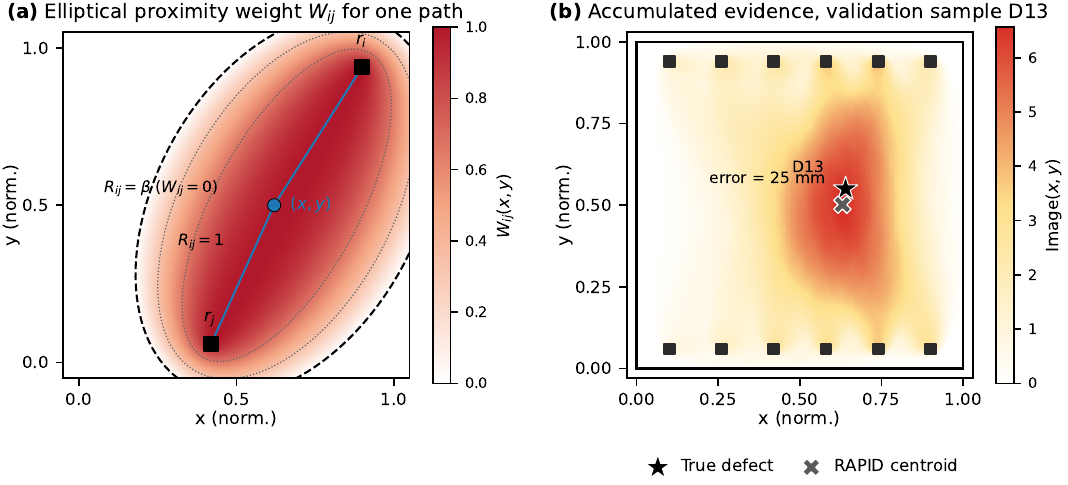}
    \caption{\textbf{RAPID's training-free imaging mechanism.} \textbf{(a)} Elliptical proximity weight $W_{ij}(x,y)$ (Eq.~\eqref{eq:rapid_weight}) for one representative propagation path between transducers $r_i$ and $r_j$: the solid contour marks $R_{ij}{=}1$ (the direct line between the two transducers), dotted contours mark intermediate values of $R_{ij}$, and the dashed outer contour marks $R_{ij}{=}\beta$, beyond which the path contributes no weight. \textbf{(b)} The accumulated evidence image $\mathrm{Image}(x,y)$ (Eq.~\eqref{eq:rapid_image}) for one Split~A validation sample (D13, near the plate center), with $\beta{=}1.05$ (the value selected by validation grid search for Split~A). Black squares mark the 12 transducers; the star marks the true defect location and the cross marks the intensity-weighted centroid prediction.}
    \label{fig:rapid_example}
\end{figure*}
\end{newpar}

Figure~\ref{fig:rapid_example}(a) illustrates the contribution of one path between sensors \(i\) and \(j\). For a candidate point \(\mathbf{q}=[x,y]^\top\), its position relative to the path is measured by
\begin{equation}
R_{ij}(\mathbf{q})
=
\frac{
\|\mathbf{q}-\mathbf{r}_i\|_2
+
\|\mathbf{q}-\mathbf{r}_j\|_2
}{
\|\mathbf{r}_i-\mathbf{r}_j\|_2
}.
\label{eq:rapid_ellipse}
\end{equation}
The value \(R_{ij}=1\) corresponds to the direct segment between the two sensors. As the candidate point moves away from this segment, \(R_{ij}\) increases along the ellipses shown in Fig.~\ref{fig:rapid_example}(a).

\new{The distance \(R_{ij}(\mathbf{q})\) only describes how far the candidate point is from path \((i,j)\). RAPID must also determine how strongly the measured deviation on that path should contribute to the image at \(\mathbf{q}\). A point close to the direct path receives a large contribution, whereas a distant point receives little or no contribution. This spatial contribution is defined by the path weight}
\begin{equation}
W_{ij}(\mathbf{q})
=
\max
\left(
0,\,
\frac{\beta-R_{ij}(\mathbf{q})}{\beta-1}
\right),
\label{eq:rapid_weight}
\end{equation}
\new{where \(\beta>1\) determines how far the influence of the path extends. The weight equals one on the direct sensor path, decreases linearly as the point moves away from it, and becomes zero at \(R_{ij}=\beta\), shown by the dashed boundary in Fig.~\ref{fig:rapid_example}(a).}

The complete RAPID image is obtained by adding the weighted contributions of all 66 measured paths:
\begin{equation}
\mathrm{Image}(\mathbf{q})
=
\sum_{(i,j)\in\mathcal{P}}
\Delta E_{ij}\,
W_{ij}(\mathbf{q}).
\label{eq:rapid_image}
\end{equation}
A path with a larger measured deviation \(\Delta E_{ij}\) contributes more strongly to the image. Regions supported by several changed paths therefore appear brighter, as shown in Fig.~\ref{fig:rapid_example}(b). The predicted location is the intensity-weighted centroid of grid points with intensity at least \(95\%\) of the image peak. If the peak is below the detection threshold, the sample is assigned the no-damage coordinate \(\mathbf{p}_{\mathrm{ud}}\).

RAPID has two parameters: the ellipse width \(\beta\) and the detection threshold. Both are selected using validation data only. 
\end{newpar}

\section{Results}
\label{sec:results}

\begin{newpar}
This section evaluates \textit{WaveGraphNet} on the OGW-1 SHM Plate benchmark under the three spatial hold-out splits defined in Section~\ref{sec:case_train_test}, presented in increasing order of extrapolation difficulty. For each split, test-set MAE is reported for the four learned baselines, the training-free RAPID baseline, standalone \textit{WaveGraphNet}, and \textit{WaveGraphNet} with test-time refinement. Results for learned models are reported as mean~\(\pm\)~standard deviation over seeds \(\{0,1,42\}\). RAPID is deterministic for a given validation-selected \(\beta\) and detection threshold and therefore has no seed variation. All errors are reported in millimeters for the \(500\,\mathrm{mm}\) plate.
\end{newpar}

\begin{newpar}
The comparison between standalone and refined \textit{WaveGraphNet} isolates the contribution of test-time refinement. Both configurations use the same inverse model and initial prediction. During refinement only the candidate coordinate is updated to improve agreement with the measured response pattern. Across the three spatial splits, this procedure reduces the mean localization error by approximately \(50\%\) relative to standalone \textit{WaveGraphNet}.
\end{newpar}

\subsection{Split A: single unseen zone}
\label{sec:results_splitA}

\begin{figure}[!htbp]
    \centering
    \includegraphics[width=1.0\linewidth]{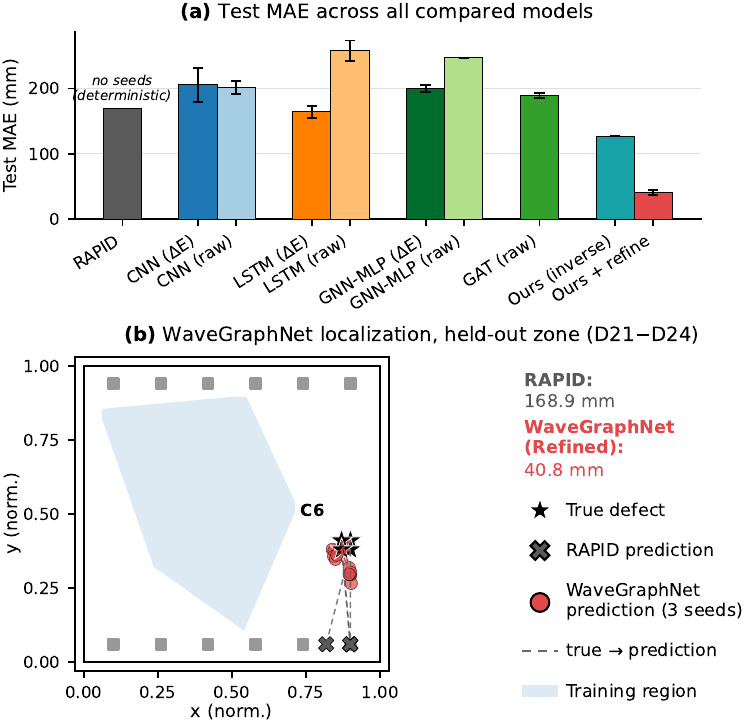}
    \caption{\textbf{Split~A results.} \textbf{(a)} Test MAE for every
    compared model (mean~$\pm$~std over 3 seeds; RAPID is deterministic).
    \textbf{(b)} \textit{WaveGraphNet} (refined) and RAPID predictions in
    the held-out zone (cluster~C6, D$_{21}$--D$_{24}$), against the
    training region's convex hull.}
    \label{fig:results_A}
\end{figure}

\begin{newpar}
Table~\ref{tab:results_splitA} reports the test-set MAE under Split~A. Standalone \textit{WaveGraphNet} achieves \(127.1\,\mathrm{mm}\), compared with \(164.0\,\mathrm{mm}\) for the best learned baseline, the LSTM with \(\Delta E\) input. Test-time refinement reduces the error further to \(40.8\,\mathrm{mm}\), corresponding to a \(67.9\%\) reduction relative to standalone \textit{WaveGraphNet}.
\end{newpar}

\begin{table}[!htbp]
\centering
\caption{Split~A test MAE. Learned models: mean~\(\pm\)~std over three seeds; RAPID: deterministic. Bold indicates the best baseline and \textit{WaveGraphNet} results.}
\label{tab:results_splitA}
\footnotesize
\setlength{\tabcolsep}{5pt}
\renewcommand{\arraystretch}{1.0}
\begin{tabular}{@{}lc@{}}
\toprule
\textbf{Model} & \textbf{Test MAE (mm)} \\
\midrule
RAPID (training-free) & 168.9 \\
\addlinespace
1D-CNN (\(\Delta E\)) & \(205.4 \pm 25.9\) \\
1D-CNN (raw) & \(201.8 \pm 9.8\) \\
LSTM (\(\Delta E\)) & \(\mathbf{164.0 \pm 9.1}\) (best baseline) \\
LSTM (raw) & \(257.6 \pm 15.6\) \\
GNN (\(\Delta E\))\textsuperscript{\(\dagger\)} & \(199.4 \pm 5.6\) \\
GNN-MLP (raw) & \(246.5 \pm 0.3\) \\
GAT (raw) & \(189.0 \pm 4.1\) \\
\addlinespace
\textbf{\textit{WaveGraphNet}, standalone} & \(\mathbf{127.1 \pm 0.4}\) \\
\textbf{\textit{WaveGraphNet}, + refinement} & \(\mathbf{40.8 \pm 3.6}\) \\
\bottomrule
\end{tabular}

\smallskip
\parbox{0.88\linewidth}{\scriptsize
\(\dagger\) With the scalar \(\Delta E\) input, GNN-MLP and GAT reduce to the same architecture; one result is therefore reported.}
\end{table}

\begin{table}[!htbp]
\centering
\caption{RAPID \(\beta\)-sweep for Split~A. Bold indicates the validation-selected value.}
\label{tab:rapid_beta_A}
\footnotesize
\setlength{\tabcolsep}{5pt}
\renewcommand{\arraystretch}{0.95}
\begin{tabular}{@{}ccc@{}}
\toprule
\(\boldsymbol{\beta}\) & \textbf{Val. MAE (mm)} & \textbf{Test MAE (mm)} \\
\midrule
1.02 & 74.3 & 167.6 \\
\textbf{1.05 (selected)} & \textbf{54.2} & 168.9 \\
1.10 & 60.8 & \textbf{68.5} \\
1.20 & 65.9 & 89.8 \\
1.30 & 66.4 & 101.1 \\
1.50 & 58.3 & 105.2 \\
2.00 & 61.3 & 120.5 \\
3.00 & 67.1 & 128.1 \\
\bottomrule
\end{tabular}
\end{table}

\begin{newpar}
The RAPID result is sensitive to the validation-selected ellipse width. As shown in Table~\ref{tab:rapid_beta_A}, \(\beta=1.05\) gives the lowest validation MAE and is therefore selected, but produces a test MAE of \(168.9\,\mathrm{mm}\). The nearby setting \(\beta=1.10\) has a validation MAE only \(6.6\,\mathrm{mm}\) higher but produces a test MAE of \(68.5\,\mathrm{mm}\). This reversal shows that RAPID's test performance on Split~A depends strongly on selecting \(\beta\) from the small validation set. The reported RAPID result correctly uses the validation-selected value \(\beta=1.05\). Even compared with the test-favorable but non-selected value \(\beta=1.10\), refined \textit{WaveGraphNet} achieves a \(40.4\%\) lower error (\(40.8\,\mathrm{mm}\) versus \(68.5\,\mathrm{mm}\) for RAPID).
\end{newpar}

\subsection{Split B: two unseen zones}
\label{sec:results_splitB}

\begin{figure}[!htbp]
    \centering
    \includegraphics[width=1.0\linewidth]{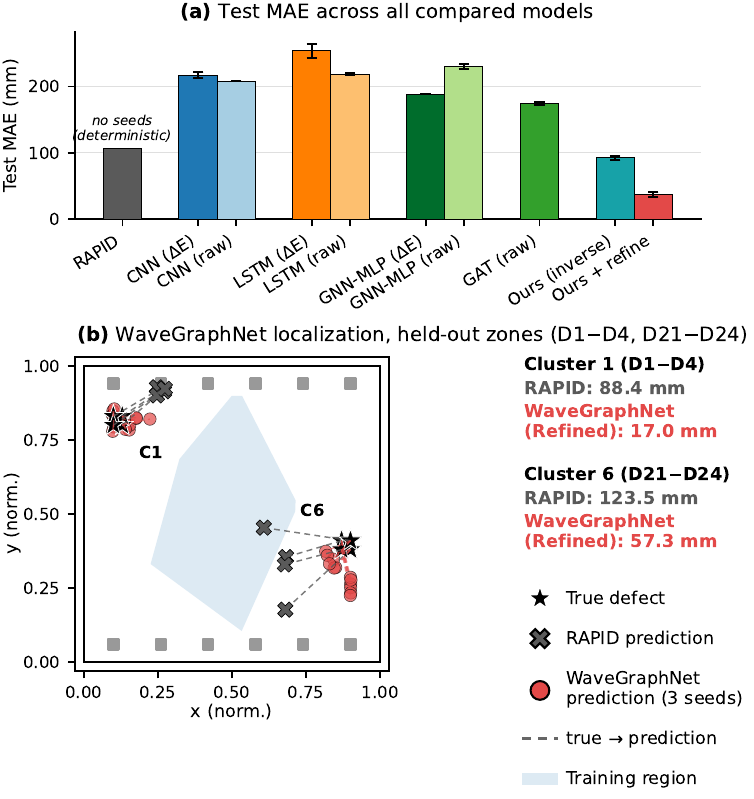}
    \caption{\textbf{Split~B results.} \textbf{(a)} Test MAE for every
    compared model. \textbf{(b)} \textit{WaveGraphNet} (refined) and RAPID
    predictions in both held-out corners (clusters~C1 and~C6), against the
    training region's convex hull.}
    \label{fig:results_B}
\end{figure}

\begin{newpar}
Split~B holds out two outer-corner clusters, D$_1$--D$_4$ and D$_{21}$--D$_{24}$, creating a harder two-region extrapolation setting than Split~A (Fig.~\ref{fig:shm_setup_case}(b)). Table~\ref{tab:results_splitB} reports the corresponding test-set MAE. RAPID is the strongest baseline at \(106.0\,\mathrm{mm}\), compared with \(174.2\,\mathrm{mm}\) for the best learned baseline, the GAT with raw-spectral input. Standalone \textit{WaveGraphNet} achieves the lowest error before refinement at \(92.5\,\mathrm{mm}\). Test-time refinement further reduces the error to \(37.2\,\mathrm{mm}\), a \(59.8\%\) improvement over standalone \textit{WaveGraphNet} and a \(64.9\%\) improvement over RAPID.
\end{newpar}

\begin{table}[!htbp]
\centering
\caption{Split~B test MAE. Learned models: mean~\(\pm\)~std over three seeds; RAPID: deterministic. Bold indicates the best baseline and \textit{WaveGraphNet} results.}
\label{tab:results_splitB}
\footnotesize
\setlength{\tabcolsep}{5pt}
\renewcommand{\arraystretch}{1.0}
\begin{tabular}{@{}lc@{}}
\toprule
\textbf{Model} & \textbf{Test MAE (mm)} \\
\midrule
RAPID (training-free) & \textbf{106.0} (best baseline) \\
\addlinespace
1D-CNN (\(\Delta E\)) & \(217.4 \pm 4.0\) \\
1D-CNN (raw) & \(208.3 \pm 0.1\) \\
LSTM (\(\Delta E\)) & \(254.1 \pm 10.5\) \\
LSTM (raw) & \(218.9 \pm 1.4\) \\
GNN (\(\Delta E\))\textsuperscript{\(\dagger\)} & \(188.7 \pm 0.6\) \\
GNN-MLP (raw) & \(230.3 \pm 3.5\) \\
GAT (raw) & \(174.2 \pm 2.2\) (best learned) \\
\addlinespace
\textbf{\textit{WaveGraphNet}, standalone} & \(\mathbf{92.5 \pm 3.4}\) \\
\textbf{\textit{WaveGraphNet}, + refinement} & \(\mathbf{37.2 \pm 3.7}\) \\
\bottomrule
\end{tabular}

\smallskip
\parbox{0.88\linewidth}{\scriptsize
\(\dagger\) GNN-MLP and GAT reduce to the same architecture with the scalar \(\Delta E\) input; one result is therefore reported.}
\end{table}

\begin{newpar}
RAPID again shows sensitivity to the ellipse width (Table~\ref{tab:rapid_beta_B}). The validation-selected value \(\beta=2.0\) gives validation and test MAEs of \(49.0\,\mathrm{mm}\) and \(106.0\,\mathrm{mm}\), respectively. The alternative \(\beta=1.1\) gives a slightly higher validation MAE of \(55.2\,\mathrm{mm}\) but a lower test MAE of \(69.6\,\mathrm{mm}\). The reported RAPID result correctly uses the validation-selected value \(\beta=2.0\). Even compared with the test-favorable but non-selected value \(\beta=1.1\), refined \textit{WaveGraphNet} achieves a \(46.6\%\) lower error (\(37.2\,\mathrm{mm}\) versus \(69.6\,\mathrm{mm}\)).
\end{newpar}

\begin{table}[!htbp]
\centering
\caption{RAPID \(\beta\)-sweep for Split~B. Bold indicates the validation-selected value and the lowest test MAE.}
\label{tab:rapid_beta_B}
\footnotesize
\setlength{\tabcolsep}{5pt}
\renewcommand{\arraystretch}{0.95}
\begin{tabular}{@{}ccc@{}}
\toprule
\(\boldsymbol{\beta}\) & \textbf{Val. MAE (mm)} & \textbf{Test MAE (mm)} \\
\midrule
1.02 & 74.5 & 115.4 \\
1.05 & 52.0 & 116.0 \\
1.10 & 55.2 & \textbf{69.6} \\
1.20 & 59.4 & 81.6 \\
1.30 & 59.8 & 90.8 \\
1.50 & 50.8 & 98.1 \\
\textbf{2.00 (selected)} & \textbf{49.0} & 106.0 \\
3.00 & 54.2 & 114.5 \\
\bottomrule
\end{tabular}
\end{table}

\begin{newpar}

Table~\ref{tab:corner_B} reports the error separately for the two test clusters. Before refinement, both RAPID and standalone \textit{WaveGraphNet} identify cluster~6 as the more difficult region: RAPID gives \(123.5\,\mathrm{mm}\) for cluster~6 and \(88.4\,\mathrm{mm}\) for cluster~1, while standalone \textit{WaveGraphNet} gives \(119.5\,\mathrm{mm}\) and \(65.6\,\mathrm{mm}\), respectively. Refinement reduces the cluster~6 error from \(119.5\,\mathrm{mm}\) to \(57.3\,\mathrm{mm}\), while the cluster~1 error decreases from \(65.6\,\mathrm{mm}\) to \(17\,\mathrm{mm}\). The larger improvement in cluster~6 shows that refinement provides its strongest correction where the initial localization error is highest.
\end{newpar}

\begin{table}[!htbp]
\centering
\caption{Split~B per-cluster MAE. RAPID uses \(n=4\), while \textit{WaveGraphNet} results are reported as mean~\(\pm\)~std over three seeds.}
\label{tab:corner_B}
\footnotesize
\setlength{\tabcolsep}{5pt}
\renewcommand{\arraystretch}{1.0}
\begin{tabular}{@{}lccc@{}}
\toprule
\textbf{Cluster} & \textbf{RAPID} & \textbf{Standalone} & \textbf{Refined} \\
\midrule
Cluster~1 (D$_1$--D$_4$) & 88.4\,mm & \(65.6 \pm 8.6\)\,mm & \(\mathbf{17.0 \pm 4.0}\)\,mm \\
Cluster~6 (D$_{21}$--D$_{24}$) & 123.5\,mm & \(119.5 \pm 2.2\)\,mm & \(\mathbf{57.3 \pm 6.3}\)\,mm \\
\bottomrule
\end{tabular}
\end{table}

\subsection{Split C: cross-region generalization}
\label{sec:results_splitC}

\begin{figure}[!htbp]
    \centering
    \includegraphics[width=1.0\linewidth]{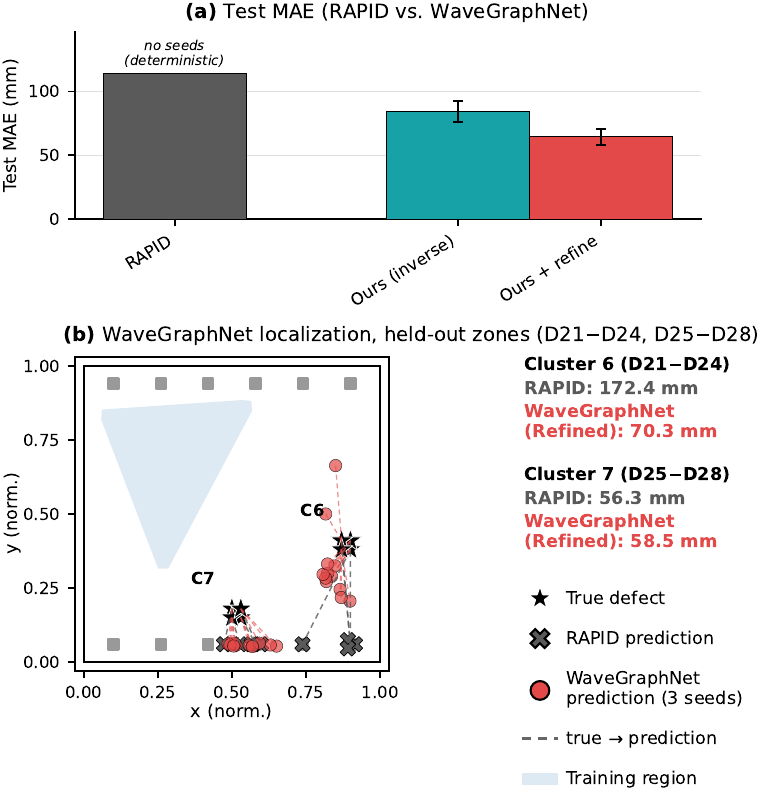}
    \caption{\textbf{Split~C results.} \textbf{(a)} Test MAE, RAPID vs.\
    \textit{WaveGraphNet}. \textbf{(b)} Predictions in both held-out
    clusters (C6 and C7), against the training region's convex hull.}
    \label{fig:results_C}
\end{figure}

\begin{newpar}
    
Split~C is the hardest setting as training is confined to one contiguous region, validation uses a spatially separate cluster, and the two test clusters lie farther beyond both regions, as shown in Fig.~\ref{fig:shm_setup_case}(c). Checkpoint selection therefore already requires extrapolation to the validation cluster before evaluation on the more distant test clusters. RAPID is the only external baseline evaluated on this split since it was the strongest baseline under Split~B. Table~\ref{tab:results_splitC} reports the test-set MAE.

Figure~\ref{fig:results_C}(a) shows that standalone \textit{WaveGraphNet} has a lower error than RAPID (\(84.3\,\mathrm{mm}\) versus \(114.4\,\mathrm{mm}\)). Test-time refinement reduces the error to \(64.4\,\mathrm{mm}\), corresponding to a \(23.6\%\) improvement over the standalone model and a \(43.7\%\) improvement over RAPID. This reduction is smaller than those obtained under Splits~A (\(67.9\%\)) and B (\(59.8\%\)), where the standalone errors leave greater room for correction.
\end{newpar}

\begin{table}[!htbp]
\centering
\caption{Split~C test MAE. \textit{WaveGraphNet}: mean~\(\pm\)~std over three seeds; RAPID: deterministic. RAPID is the only baseline evaluated on this split.}
\label{tab:results_splitC}
\footnotesize
\setlength{\tabcolsep}{5pt}
\renewcommand{\arraystretch}{1.0}
\begin{tabular}{@{}lc@{}}
\toprule
\textbf{Model} & \textbf{Test MAE (mm)} \\
\midrule
RAPID (training-free) & \textbf{114.4} \\
\textit{WaveGraphNet}, standalone & \(84.3 \pm 7.9\) \\
\textbf{\textit{WaveGraphNet}, + refinement} & \(\mathbf{64.4 \pm 6.2}\) \\
\bottomrule
\end{tabular}
\end{table}

\begin{newpar}

RAPID remains sensitive to the ellipse width (Table~\ref{tab:rapid_beta_C}). The validation-selected value \(\beta=1.05\) gives validation and test MAEs of \(16.8\,\mathrm{mm}\) and \(114.4\,\mathrm{mm}\), respectively. The alternative \(\beta=1.10\) gives a slightly higher validation MAE of \(20.0\,\mathrm{mm}\) but a substantially lower test MAE of \(66.3\,\mathrm{mm}\). Although this alternative is not used in the primary comparison because it was not selected using validation data, its test error remains slightly higher than that of refined \textit{WaveGraphNet} (\(64.4\,\mathrm{mm}\)).

Figure~\ref{fig:results_C}(b) and Table~\ref{tab:corner_C} show the cluster-wise spatial localization performance, with cluster~6 being more difficult than cluster~7 before refinement. Refinement reduces the \textit{WaveGraphNet} error on cluster~6 from \(131.8\,\mathrm{mm}\) to \(70.3\,\mathrm{mm}\), but increases the error on cluster~7 from \(36.7\,\mathrm{mm}\) to \(58.5\,\mathrm{mm}\). The aggregate improvement on Split~C therefore results from the large correction on the more difficult cluster~6, which outweighs the performance degradation on cluster~7.
\end{newpar}

\begin{table}[!htbp]
\centering
\caption{RAPID \(\beta\)-sweep for Split~C. Bold indicates the validation-selected value and the lowest test MAE.}
\label{tab:rapid_beta_C}
\footnotesize
\setlength{\tabcolsep}{5pt}
\renewcommand{\arraystretch}{0.95}
\begin{tabular}{@{}ccc@{}}
\toprule
\(\boldsymbol{\beta}\) & \textbf{Val. MAE (mm)} & \textbf{Test MAE (mm)} \\
\midrule
1.02 & 131.2 & 113.6 \\
\textbf{1.05 (selected)} & \textbf{16.8} & 114.4 \\
1.10 & 20.0 & \textbf{66.3} \\
1.20 & 20.3 & 74.6 \\
1.30 & 22.1 & 77.6 \\
1.50 & 24.9 & 81.5 \\
2.00 & 25.0 & 100.3 \\
3.00 & 24.3 & 101.1 \\
\bottomrule
\end{tabular}
\end{table}

\begin{table}[!htbp]
\centering
\caption{Split~C per-cluster MAE (mm). \textit{WaveGraphNet} results are reported as mean~\(\pm\)~std over three seeds.}
\label{tab:corner_C}
\footnotesize
\setlength{\tabcolsep}{5pt}
\renewcommand{\arraystretch}{1.0}
\begin{tabular}{@{}lccc@{}}
\toprule
\textbf{Cluster} & \textbf{RAPID} & \textbf{Standalone} & \textbf{Refined} \\
\midrule
Cluster~6 (D$_{21}$--D$_{24}$) & 172.4 & \(131.8 \pm 3.9\) & \(\mathbf{70.3 \pm 12.1}\) \\
Cluster~7 (D$_{25}$--D$_{28}$) & 56.3 & \(36.7 \pm 11.9\) & \(58.5 \pm 1.8\) \\
\bottomrule
\end{tabular}
\end{table}

\subsection{Cross-split synthesis}
\label{sec:results_synthesis}

\begin{newpar}
Across the three validation-selected comparisons, standalone \textit{WaveGraphNet} achieves the lowest reported test MAE, and refinement reduces its error by an average of \(50.4\%\). 

The baseline ranking varies across Splits~A and B. RAPID is the strongest baseline on Split~B and remains competitive in the Split~A parameter sweep; it was therefore retained as the external baseline for Split~C. However, RAPID is sensitive to the selected ellipse width: competing \(\beta\) values differ by only \(3.2\)--\(6.6\,\mathrm{mm}\) on validation but by \(36.4\)--\(100.4\,\mathrm{mm}\) on test. All primary comparisons use the validation-selected value.
\end{newpar}

\subsection{False-positive rate on undamaged samples}
\label{sec:results_fpr}

\begin{newpar}
The false-positive rate (FPR) is the fraction of held-out undamaged samples whose predicted coordinates fall inside the plate domain \(\Omega=[0,1]^2\). During training, undamaged samples are assigned the target coordinate \(\mathbf{p}_{\mathrm{ud}}=(-0.5,-0.5)\), located outside this domain. For \textit{WaveGraphNet}, FPR is reported for the standalone model because test-time refinement is applied only to damaged samples and therefore does not affect undamaged predictions. Table~\ref{tab:fpr_all} reports FPR together with test MAE for each model and split. All evaluated models achieve \(0.0\%\) FPR, meaning that none of the held-out undamaged samples is incorrectly localized inside the plate. Thus, every model successfully recovers the no-damage behavior under the present evaluation protocol.
\end{newpar}

\begin{table}[!h]
\centering
\caption{Test MAE on damaged samples and FPR on held-out undamaged samples. Learned-model FPR uses 18 evaluations per split; RAPID is deterministic. Refinement is excluded because it is applied only to damaged samples.}
\label{tab:fpr_all}
\scriptsize
\setlength{\tabcolsep}{4pt}
\renewcommand{\arraystretch}{0.9}
\begin{tabular}{@{}lcc@{}}
\toprule
\textbf{Model} & \textbf{Test MAE (mm)} & \textbf{FPR} \\
\midrule
\multicolumn{3}{@{}l}{\textit{Split~A}} \\
RAPID & 168.9 & \textbf{0.0\%} \\
1D-CNN (raw) & 201.8 & \textbf{0.0\%} \\
LSTM (raw) & 257.6 & \textbf{0.0\%} \\
GNN-MLP (raw) & 246.5 & \textbf{0.0\%} \\
GAT (raw) & 189.0 & \textbf{0.0\%} \\
1D-CNN (\(\Delta E\)) & 205.4 & \textbf{0.0\%} \\
LSTM (\(\Delta E\)) & 164.0 & \textbf{0.0\%} \\
GNN (\(\Delta E\)) & 199.4 & \textbf{0.0\%} \\
\textbf{\textit{WaveGraphNet} (standalone)} & \textbf{127.1} & \textbf{0.0\%} \\
\midrule
\multicolumn{3}{@{}l}{\textit{Split~B}} \\
RAPID & 106.0 & \textbf{0.0\%} \\
1D-CNN (raw) & 208.3 & \textbf{0.0\%} \\
LSTM (raw) & 218.9 & \textbf{0.0\%} \\
GNN-MLP (raw) & 230.3 & \textbf{0.0\%} \\
GAT (raw) & 174.2 & \textbf{0.0\%} \\
1D-CNN (\(\Delta E\)) & 217.4 & \textbf{0.0\%} \\
LSTM (\(\Delta E\)) & 254.1 & \textbf{0.0\%} \\
GNN (\(\Delta E\)) & 188.7 & \textbf{0.0\%} \\
\textbf{\textit{WaveGraphNet} (standalone)} & \textbf{92.5} & \textbf{0.0\%} \\
\midrule
\multicolumn{3}{@{}l}{\textit{Split~C}} \\
RAPID & \textbf{114.4} & \textbf{0.0\%} \\
\textit{WaveGraphNet} (standalone) & 84.3 & \textbf{0.0\%} \\
\bottomrule
\end{tabular}
\end{table}

All models achieve \(0.0\%\) FPR on every evaluated split. Undamaged predictions therefore remain outside the plate domain across all models, splits, and seeds. Test-time refinement is not applied to undamaged samples and consequently does not alter these results.

\section{Conclusion}
\label{sec:discussion}

\begin{newpar}
This paper presented \textit{WaveGraphNet}, an inverse--forward graph framework for guided-wave damage localization in CFRP plates. The inverse branch maps the measured path-wise response deviations to a defect coordinate, while the independently trained forward branch predicts the response pattern associated with a candidate coordinate. At test time, the gradients through the forward model refines the inverse prediction by updating only the candidate coordinate to improve agreement with the measured response.

The framework was evaluated on the OGW-1 SHM Plate benchmark using three spatial hold-out splits, with the training-free RAPID imaging method as a classical reference. RAPID outperformed all learned baselines on Split~B, while its parameter sweep also showed competitive performance on Split~A, confirming that it provides a strong baseline. Standalone and refined \textit{WaveGraphNet} achieved the lowest reported test MAE under the validation-selected settings on all three splits. Refinement reduced the standalone error by \(23.6\%\)--\(67.9\%\), with an average reduction of \(50.4\%\). All evaluated models achieved a \(0.0\%\) false-positive rate on undamaged samples. The central finding is that gradients from a learned forward response model can improve spatial extrapolation by refining the predicted damage coordinate according to response consistency.

The cluster-level results on Split~C, however, show that refinement does not improve every prediction: the substantial correction in the more difficult cluster is accompanied by an increase in error in the other cluster, where the initial localization was already comparatively accurate. This motivates uncertainty-aware refinement that quantifies confidence in the predicted forward response and uses it to control the magnitude or acceptance of coordinate updates, preserving the initial inverse prediction when the evidence for correction is weak, in the spirit of uncertainty-aware and out-of-distribution-aware diagnostics~\cite{han2022ood,basora2025uq}. Future work should also integrate refinement with damage detection and evaluate both localization accuracy and uncertainty calibration across different specimens, sensing layouts, and environmental conditions.

\end{newpar}

\vspace{0.05\linewidth}
\noindent\textbf{Code and Data Availability:}
The \textit{WaveGraphNet} implementation is publicly available at
\url{https://github.com/vinaySHARMA-mech/WaveGraphNet}, including preprocessing scripts, graph-construction code,
training configurations, and the spatial hold-out split definitions used in this
study. The released splits include Split A, Split B, and Split C for the OGW-1 SHM
Plate benchmark. The trained model checkpoints for all three splits and all three seeds are
archived at \url{https://doi.org/10.5281/zenodo.22639832}, and allow the reported
results to be reproduced without retraining. The raw OGW-1 dataset is publicly available from the original
benchmark source \cite{ogw1}.

\vspace{0.05\linewidth}
\noindent\textbf{Acknowledgment:}
This research was funded by the Swiss National Science Foundation (SNSF) Grant Number 200021\_200461.

\vspace{0.05\linewidth}
\noindent\textbf{Declaration of Generative AI and AI-assisted technologies in the writing process:}
During the preparation of this work, the authors used ChatGPT to assist with refining and correcting the text. After using this tool, the authors carefully reviewed and edited the content as needed and take full responsibility for the content of this publication.
\bibliographystyle{elsarticle-num}
\bibliography{cas-refs}

@article{Wang2022GiG,
  author={Wang, Sheng and Luo, Zhitao and Shen, Peng and Zhang, Hui and Ni, Zhonghua},
  title={Graph-in-Graph Convolutional Network for Ultrasonic Guided Wave-Based Damage Detection and Localization},
  journal={IEEE Transactions on Instrumentation and Measurement},
  year={2022},
  volume={71},
  pages={1--11},
  doi={10.1109/TIM.2022.3144732}
}

@article{sun2024physics_augmented_stgcn_guided_waves,
  author={Lingyu Sun and Ruijie Song and Juntao Wei and Yumeng Gao and Chang Peng and Longqing Fan and Mingshun Jiang and Lei Zhang},
  title={Physics-Augmented Spatial-Temporal graph convolutional network for damage localization using Ultrasonic guided waves},
  journal={Mechanical Systems and Signal Processing},
  volume={221},
  pages={111738},
  year={2024},
  doi={10.1016/j.ymssp.2024.111738}
}

@article{guemes2020composite_shm_review,
  author={Güemes, Alfredo and Fernandez-Lopez, Antolin and Pozo, Antonio Rodriguez and Sierra-Perez, Jorge},
  title={Structural Health Monitoring for Advanced Composite Structures: A Review},
  journal={Journal of Composites Science},
  volume={4},
  number={1},
  pages={13},
  year={2020},
  doi={10.3390/jcs4010013}
}

@article{cawley2018closing_gap,
  author={Cawley, P.},
  title={Structural health monitoring: Closing the gap between research and industrial deployment},
  journal={Structural Health Monitoring},
  volume={17},
  number={5},
  pages={1225--1244},
  year={2018},
  doi={10.1177/1475921717750047}
}

@article{ricci2022guided_waves_composites_review,
  author={Ricci, Fabrizio and Monaco, E. and Boffa, N. D. and Maio, L. and Memmolo, V.},
  title={Guided waves for structural health monitoring in composites: A review and implementation strategies},
  journal={Progress in Aerospace Sciences},
  volume={129},
  pages={100790},
  year={2022},
  doi={10.1016/j.paerosci.2021.100790}
}

@article{capineri2021guided_wave_sensors_review,
  author={Capineri, Lorenzo and Bulletti, Andrea},
  title={Ultrasonic Guided-Waves Sensors and Integrated Structural Health Monitoring Systems for Impact Detection and Localization: A Review},
  journal={Sensors},
  volume={21},
  number={9},
  pages={2929},
  year={2021},
  doi={10.3390/s21092929}
}

@article{azad2024lamb_wave_localization_severity,
  author={Azad, Muhammad Muzammil and Munyaneza, Olivier and Jung, Jaehyun and Sohn, Jung Woo and Han, Jang-Woo and Kim, Heung Soo},
  title={Damage Localization and Severity Assessment in Composite Structures Using Deep Learning Based on Lamb Waves},
  journal={Sensors},
  volume={24},
  number={24},
  pages={8057},
  year={2024},
  doi={10.3390/s24248057}
}

@article{rautela2021model_assisted_guided_wave,
  author={Rautela, Mahindra and Gopalakrishnan, S.},
  title={Ultrasonic guided wave based structural damage detection and localization using model assisted convolutional and recurrent neural networks},
  journal={Expert Systems with Applications},
  volume={167},
  pages={114189},
  year={2021},
  doi={10.1016/j.eswa.2020.114189}
}

@article{battaglia2018relational_inductive_biases,
  author={Battaglia, Peter W. and Hamrick, Jessica B. and Bapst, Victor and Sanchez-Gonzalez, Alvaro and Zambaldi, Vinicius and Malinowski, Mateusz and Tacchetti, Andrea and Raposo, David and Santoro, Adam and Faulkner, Ryan and Gulcehre, Caglar and Song, Francis and Ballard, Andrew and Gilmer, Justin and Dahl, George E. and Vaswani, Ashish and Allen, Kelsey R. and Nash, Charles and Langston, Victoria and Dyer, Chris and Heess, Nicolas and Wierstra, Daan and Kohli, Pushmeet and Botvinick, Matthew and Vinyals, Oriol and Li, Yujia and Pascanu, Razvan},
  title={Relational inductive biases, deep learning, and graph networks},
  journal={arXiv preprint arXiv:1806.01261},
  year={2018}
}

@inproceedings{velickovic2018gat,
  author={Veli\v{c}kovi\'{c}, Petar and Cucurull, Guillem and Casanova, Arantxa and Romero, Adriana and Li\`o, Pietro and Bengio, Yoshua},
  title={Graph Attention Networks},
  booktitle={International Conference on Learning Representations},
    url={https://openreview.net/forum?id=rJXMpikCZ},

  year={2018}
}

@article{ogw1,
  author={Moll, Jochen and Kathol, Jens and Fritzen, Claus-Peter and Moix-Bonet, Maria and Rennoch, Marcel and Koerdt, Michael and Herrmann, Axel S. and Sause, Markus G. R. and Bach, Martin},
  title={Open Guided Waves: online platform for ultrasonic guided wave measurements},
  journal={Structural Health Monitoring},
  volume={18},
  number={5--6},
  pages={1903--1914},
  year={2019},
  doi={10.1177/1475921718817169}
}

@article{cawley2024guided_waves_ndt_shm,
  author={Cawley, Peter},
  title={Guided waves in long range nondestructive testing and structural health monitoring: Principles, history of applications and prospects},
  journal={NDT \& E International},
  volume={142},
  pages={103026},
  year={2024},
  doi={10.1016/j.ndteint.2023.103026}
}

@article{tanveer2024guided_waves_laminated_composites,
  author={Tanveer, Mohad and Elahi, Muhammad Umar and Jung, Jaehyun and Azad, Muhammad Muzammil and Khalid, Salman and Kim, Heung Soo},
  title={Recent Advancements in Guided Ultrasonic Waves for Structural Health Monitoring of Composite Structures},
  journal={Applied Sciences},
  volume={14},
  number={23},
  pages={11091},
  year={2024},
  doi={10.3390/app142311091}
}

@article{philibert2022lamb_waves_aeronautics,
  author={Philibert, Marilyne and Yao, Kui and Gresil, Matthieu and Soutis, Constantinos},
  title={Lamb waves-based technologies for structural health monitoring of composite structures for aircraft applications},
  journal={European Journal of Materials},
  volume={2},
  number={1},
  pages={436--474},
  year={2022},
  doi={10.1080/26889277.2022.2094839}
}

@article{sattarifar2022ml_guided_wave_shm_review,
  author={Sattarifar, Afshin and Nestorovi\'{c}, Tamara},
  title={Emergence of Machine Learning Techniques in Ultrasonic Guided Wave-based Structural Health Monitoring},
  journal={International Journal of Prognostics and Health Management},
  volume={13},
  number={1},
  year={2022},
  doi={10.36001/ijphm.2022.v13i1.3107}
}

@article{liao2023gramian_guided_wave_localization,
  author={Liao, Yunlai and Qing, Xinlin and Wang, Yihan and Zhang, Fanghong},
  title={Damage localization for composite structure using guided wave signals with Gramian angular field image coding and convolutional neural networks},
  journal={Composite Structures},
  volume={312},
  pages={116871},
  year={2023},
  doi={10.1016/j.compstruct.2023.116871}
}

@article{gao2024lamb_wave_modular_ann,
  author={Gao, Yumeng and Sun, Lingyu and Song, Ruijie and Peng, Chang and Wu, Xiaobo and Wei, Juntao and Jiang, Mingshun and Sui, Qingmei and Zhang, Lei},
  title={Damage localization in composite structures based on Lamb wave and modular artificial neural network},
  journal={Sensors and Actuators A: Physical},
  volume={377},
  pages={115644},
  year={2024},
  doi={10.1016/j.sna.2024.115644}
}

@article{lomazzi2023unified_damage_detection_localization_quantification,
  author={Lomazzi, Luca and Giglio, Marco and Cadini, Francesco},
  title={Towards a deep learning-based unified approach for structural damage detection, localisation and quantification},
  journal={Engineering Applications of Artificial Intelligence},
  volume={121},
  pages={106003},
  year={2023},
  doi={10.1016/j.engappai.2023.106003}
}

@article{bloemheuvel2021graph_shm_framework,
  author={Bloemheuvel, Stefan and van den Hoogen, Jurgen and Atzmueller, Martin},
  title={A computational framework for modeling complex sensor network data using graph signal processing and graph neural networks in structural health monitoring},
  journal={Applied Network Science},
  volume={6},
  number={1},
  pages={97},
  year={2021},
  doi={10.1007/s41109-021-00438-8}
}

@article{cheema2024graph_signal_processing_shm,
  author={Cheema, Muhammad Asaad and Sarwar, Muhammad Zohaib and Gogineni, Vinay Chakravarthi and Cantero, Daniel and Rossi, Pierluigi Salvo},
  title={Computationally Efficient Structural Health Monitoring Using Graph Signal Processing},
  journal={IEEE Sensors Journal},
  volume={24},
  number={7},
  pages={11895--11905},
  year={2024},
  doi={10.1109/JSEN.2024.3366346}
}

@article{ongie2020inverse_imaging_review,
  author={Ongie, Gregory and Jalal, Ajil and Metzler, Christopher A. and Baraniuk, Richard G. and Dimakis, Alexandros G. and Willett, Rebecca},
  title={Deep Learning Techniques for Inverse Problems in Imaging},
  journal={IEEE Journal on Selected Areas in Information Theory},
  volume={1},
  number={1},
  pages={39--56},
  year={2020},
  doi={10.1109/JSAIT.2020.2991563}
}

@article{raissi2019physics,
  author={Raissi, M. and Perdikaris, P. and Karniadakis, G. E.},
  title={Physics-informed neural networks: A deep learning framework for solving forward and inverse problems involving nonlinear partial differential equations},
  journal={Journal of Computational Physics},
  volume={378},
  pages={686--707},
  year={2019},
  doi={10.1016/j.jcp.2018.10.045}
}

@article{song2024physics_guided_lamb_wave,
  author={Song, Yang and Shan, Shengbo and Zhang, Yuanman and Cheng, Li},
  title={Physics-guided neural network for structural health monitoring with lamb waves through boundary reflection elimination},
  journal={Structural Health Monitoring},
  volume={25},
  number={2},
  pages={1219--1236},
  year={2026},
  doi={10.1177/14759217241305050}
}

@article{delpriore2025gnn_aerospace_damage_localization,
  author={Del Priore, Emiliano and Lampani, Luca},
  title={Real-Time Damage Detection and Localization on Aerospace Structures Using Graph Neural Networks},
  journal={Journal of Sensor and Actuator Networks},
  volume={14},
  number={5},
  pages={89},
  year={2025},
  doi={10.3390/jsan14050089}
}

@article{fink2026physics,
  title={From physics to machine learning and back: {Part I} -- Learning with inductive biases in prognostics and health management ({PHM})},
  author={Fink, Olga and Sharma, Vinay and Nejjar, Ismail and Von Krannichfeldt, Leandro and Garmaev, Sergei and Zhang, Zepeng and Wei, Amaury and Frusque, Gaetan and Forest, Florent and Zhao, Mengjie and Hsu, Chi-Ching and {Faghih Niresi}, Keivan and Sun, Han and Dong, Hao and Xu, Chenghao and Theiler, Raffael and Bizzi, Arthur and Steiner, Kevin},
  journal={Reliability Engineering \& System Safety},
  volume={271},
  pages={112213},
  year={2026},
  publisher={Elsevier},
  doi={10.1016/j.ress.2026.112213}
}

@article{fink2026physics2,
  title={From physics to machine learning and back: {Part II} -- Learning and observational bias in prognostics and health management ({PHM})},
  author={Fink, Olga and Nejjar, Ismail and Sharma, Vinay and {Faghih Niresi}, Keivan and Sun, Han and Dong, Hao and Xu, Chenghao and Wei, Amaury and Bizzi, Arthur and Theiler, Raffael and Tian, Yuan and Von Krannichfeldt, Leandro and Ma, Zhan and Garmaev, Sergei and Zhang, Zepeng and Zhao, Mengjie and Steiner, Kevin and Kesmen, Yusuf},
  journal={Reliability Engineering \& System Safety},
  volume={274},
  pages={112376},
  year={2026},
  publisher={Elsevier},
  doi={10.1016/j.ress.2026.112376}
}

@article{niresi2025rins,
  title={RINS-T: Robust Implicit Neural Solvers for Time-Series Linear Inverse Problems},
  author={Niresi, Keivan Faghih and Zhang, Zepeng and Fink, Olga},
  journal={IEEE Transactions on Instrumentation and Measurement},
  volume={74},
  pages={1--14},
  year={2025},
  publisher={IEEE},
  doi={10.1109/TIM.2025.3632427}
}

@article{song2025unsupervised,
  title={Unsupervised temperature-compensated damage localization method based on damage to baseline autoencoder and delay-based probabilistic imaging},
  author={Song, Ruijie and Sun, Lingyu and Gao, Yumeng and Wei, Juntao and Peng, Chang and Fan, Longqing and Jiang, Mingshun},
  journal={Mechanical Systems and Signal Processing},
  volume={230},
  pages={112649},
  year={2025},
  publisher={Elsevier},
  doi={10.1016/j.ymssp.2025.112649}
}

@inproceedings{wang2017time,
  title={Time series classification from scratch with deep neural networks: A strong baseline},
  author={Wang, Zhiguang and Yan, Weizhong and Oates, Tim},
  booktitle={2017 International joint conference on neural networks (IJCNN)},
  pages={1578--1585},
  year={2017},
  organization={IEEE}
}

@article{hochreiter1997long,
  author={Hochreiter, Sepp and Schmidhuber, J\"urgen},
  title={Long Short-Term Memory},
  journal={Neural Computation},
  volume={9},
  number={8},
  pages={1735--1780},
  year={1997},
  doi={10.1162/neco.1997.9.8.1735}
}

@inproceedings{gao2005rapid,
  author    = {Gao, H. and Shi, Y. and Rose, Joseph L.},
  title     = {Guided Wave Tomography on an Aircraft Wing with Leave in Place Sensors},
  booktitle = {Review of Progress in Quantitative Nondestructive Evaluation, Volume 24},
  series    = {AIP Conference Proceedings},
  volume    = {760},
  pages     = {1788--1794},
  year      = {2005},
  publisher = {AIP},
  doi       = {10.1063/1.1916887},
  url       = {https://doi.org/10.1063/1.1916887}
}

@article{zhao2007rapid,
  author  = {Zhao, Xiaoliang and Gao, Huidong and Zhang, Guangfan and
             Ayhan, Bulent and Yan, Fei and Kwan, Chiman and Rose, Joseph L.},
  title   = {Active Health Monitoring of an Aircraft Wing with Embedded
             Piezoelectric Sensor/Actuator Network: I. Defect Detection,
             Localization and Growth Monitoring},
  journal = {Smart Materials and Structures},
  volume  = {16},
  number  = {4},
  pages   = {1208--1217},
  year    = {2007},
  doi     = {10.1088/0964-1726/16/4/032},
  url     = {https://doi.org/10.1088/0964-1726/16/4/032}
}

@article{vanvelsor2007rapid,
  author    = {{Van Velsor}, Jason K. and Gao, Huidong and Rose, Joseph L.},
  title     = {Guided-Wave Tomographic Imaging of Defects in Pipe Using a
               Probabilistic Reconstruction Algorithm},
  journal   = {Insight: Non-Destructive Testing and Condition Monitoring},
  volume    = {49},
  number    = {9},
  pages     = {532--537},
  year      = {2007},
  publisher = {British Institute of Non-Destructive Testing},
  doi       = {10.1784/insi.2007.49.9.532},
  url       = {https://doi.org/10.1784/insi.2007.49.9.532}
}

@article{hettler2016rapid,
  author  = {Hettler, Jan and Tabatabaeipour, Morteza and Delrue, Steven and
             {Van Den Abeele}, Koen},
  title   = {Linear and Nonlinear Guided Wave Imaging of Impact Damage in
             {CFRP} Using a Probabilistic Approach},
  journal = {Materials},
  volume  = {9},
  number  = {11},
  pages   = {901},
  year    = {2016},
  doi     = {10.3390/ma9110901},
  url     = {https://doi.org/10.3390/ma9110901}
}

@inproceedings{zaheer2017deep,
  author    = {Zaheer, Manzil and Kottur, Satwik and Ravanbakhsh, Siamak and Poczos, Barnabas and Salakhutdinov, Ruslan and Smola, Alexander J.},
  title     = {Deep Sets},
  booktitle = {Advances in Neural Information Processing Systems (NeurIPS)},
  year      = {2017},
}

@article{niresi2026timevertex,
  author={Faghih Niresi, Keivan and Qing, Jun and Zhao, Mengjie and Fink, Olga},
  title={Time-Vertex machine learning for optimal sensor placement in temporal graph signals: Applications in structural health monitoring},
  journal={Reliability Engineering \& System Safety}, volume={270}, pages={112153}, year={2026},
  doi={10.1016/j.ress.2025.112153}}

@article{liu2024structural,
  author={Liu, Jie and Li, Qilin and Li, Ling and An, Senjian},
  title={Structural damage detection and localization via an unsupervised anomaly detection method},
  journal={Reliability Engineering \& System Safety}, volume={252}, pages={110465}, year={2024},
  doi={10.1016/j.ress.2024.110465}}

@article{miele2023multifidelity,
  author={Miele, S. and Karve, P. and Mahadevan, S.},
  title={Multi-fidelity physics-informed machine learning for probabilistic damage diagnosis},
  journal={Reliability Engineering \& System Safety}, volume={235}, pages={109243}, year={2023},
  doi={10.1016/j.ress.2023.109243}}

@article{han2022ood,
  author={Han, Te and Li, Yan-Fu},
  title={Out-of-distribution detection-assisted trustworthy machinery fault diagnosis approach with uncertainty-aware deep ensembles},
  journal={Reliability Engineering \& System Safety}, volume={226}, pages={108648}, year={2022},
  doi={10.1016/j.ress.2022.108648}}

@article{basora2025uq,
  author={Basora, Luis and Viens, Arthur and Chao, Manuel Arias and Olive, Xavier},
  title={A benchmark on uncertainty quantification for deep learning prognostics},
  journal={Reliability Engineering \& System Safety}, volume={253}, pages={110513}, year={2025},
  doi={10.1016/j.ress.2024.110513}}

\end{document}